\documentclass[graybox, envcountchap]{svmult}
\def\nofake{true}

\PassOptionsToPackage{dvipsnames}{xcolor}

\usepackage{imakeidx}

\usepackage[utf8]{inputenc}
\inputencoding{utf8}

\usepackage{xstring}
\usepackage{xparse}
\usepackage{expl3}

\usepackage{bookmark}
\bookmarksetup{
  numbered,
}
\makeatletter
\providecommand*{\toclevel@titlech}{0}
\edef\toclevel@authorch{\the\numexpr\toclevel@titlech+1}
\makeatother

\makeatletter
\newif\ifallcaps
\newcommand{\checkcaps}[1]{\edef\@tempa{\lowercase{#1}}\ifx\@tempa#1\allcapstrue
  \else
    \allcapsfalse
  \fi
}
\makeatother

\makeatletter
\newcommand*{\MyIndex}[1]{\lowercase{\def\temp{#1}}\uppercase{\def\utemp{#1}}\def\otemp{#1}\ifx\utemp\otemp \expandafter\index\expandafter{\utemp}\else \expandafter\index\expandafter{\temp}\fi }\makeatother

\usepackage{makeidx}         \makeindex                   \ifdefined\nofake
  \usepackage{graphicx}      \else
  \usepackage[demo]{graphicx}\fi
\usepackage[bottom]{footmisc}\usepackage{multicol}        \usepackage{multirow}

\usepackage{mathtools}
\usepackage[detect-weight=true]{siunitx}
\usepackage{chapterbib}
\usepackage{subcaption}
\usepackage{booktabs}
\usepackage{bold-extra}
\usepackage[dvipsnames]{xcolor}
\usepackage{listings}
\usepackage{cleveref}
\usepackage{enumerate}

\usepackage{mathptmx}       \usepackage{helvet}         \usepackage{courier}        \usepackage{type1cm}        

\usepackage{amsmath}
\usepackage{cite}
\interdisplaylinepenalty=2500 \usepackage{url}
\usepackage{tikz}             \usepackage{tikz-qtree}
\usetikzlibrary{angles, arrows, calc, quotes, matrix, cd, positioning, arrows.meta, fit, pgfplots.groupplots}
\usepackage{float}
\usepackage{algorithm}
\usepackage{algpseudocode}
\usepackage[strings]{underscore}

\usepackage[acronym,shortcuts]{glossaries}
\newglossary[algh]{hidden}{acrh}{acnh}{Hidden Acronyms}

\newacronym{ML}{ML}{Machine Learning}
\newacronym{EC}{EC}{Evolutionary Computation}
\newacronym{GP}{GP}{Genetic Programming}

\definecolor{codegreen}{rgb}{0,0.6,0}
\definecolor{codegray}{rgb}{0.5,0.5,0.5}
\definecolor{codepurple}{rgb}{0.58,0,0.82}
\definecolor{backcolour}{rgb}{0.95,0.95,0.92}

\lstdefinestyle{mystyle}{
  language=python,
  tabsize=3,
  label=code:sample,
  frame=shadowbox,
  rulesepcolor=\color{codegray},
  xleftmargin=20pt,
  framexleftmargin=15pt,
  keywordstyle=\color{blue}\bf,
  commentstyle=\color{codegreen},
  stringstyle=\color{red},
  numbers=left,
  numberstyle=\tiny,
  numbersep=5pt,
  breaklines=true,
  showstringspaces=false,
  basicstyle=\footnotesize,
  emph={str},
  emphstyle={\color{magenta}},
  showtabs=false
}
\usepackage{colortbl}

\usepackage{etoolbox}
\AtBeginDocument{\patchcmd{\bibsection}{\addcontentsline{toc}{section}{\refname}}{}{}{}}

\usepackage{import}

\begin{document}

\title{Trust, but Verify: Rigorously Profiling Best-Effort High-Performance Computing for Digital Evolution}
\titlerunning{Trust, but Verify: Rigorously Profiling Best-Effort HPC}

\author{Matthew Andres Moreno, Santiago Rodriguez Papa, Charles Ofria, Luis Zaman, and Emily Dolson}
\authorrunning{M.A. Moreno et al.}

\institute{MAM{\scriptsize\,\href{https://orcid.org/0000-0003-4726-4479}{(ORCID\,0000-0003-4726-4479)}}, LHZ{\scriptsize\,\href{https://orcid.org/0000-0001-6838-7385}{(ORCID\,0000-0001-6838-7385)}} \at
Department of Ecology and Evolutionary Biology;
Center for the Study of Complex Systems;
University of Michigan, Ann Arbor, MI, USA,
\email{morenoma@umich.edu}
\\[3pt]
SRP{\scriptsize\,\href{https://orcid.org/0000-0002-6028-2105}{(ORCID\,0000-0002-6028-2105)}}, CAO{\scriptsize\,\href{https://orcid.org/0000-0003-2924-1732}{(ORCID\,0000-0003-2924-1732)}}, ELD{\scriptsize\,\href{https://orcid.org/0000-0001-8616-4898}{(ORCID\,0000-0001-8616-4898)}} \at
Department of Computer Science and Engineering;
Program in Ecology, Evolution, and Behavior;
Michigan State University, East Lansing, MI, USA}

\maketitle
 
\abstract{Developments in high-performance computing (HPC) technology continue to drastically increase quantities of available processing power.
In the context of digital evolution, this explosive growth offers opportunities to advance both hypothesis-driven explorations of multi-scale biological phenomena and application-driven evolutionary optimization targeting hard problem domains.
A particular opportunity arises from emerging next-generation AI/ML hardware accelerator platforms, such as the 880,000-processor Cerebras Wafer-Scale Engine (WSE).
Such hardware, however, constrains on-device data storage and movement --- a challenge compounded by vulnerability to failures arising over numerous device components.
Best-effort relaxations that depart from a traditional deterministic computing paradigm can help accommodate such constraints, but complicate reproducibility and risk introducing artifactual biases.
We explore these concerns, developing a framework to measure runtime behavior of best-effort code and examining case studies of best-effort computing in digital evolution projects.
The first case study applies best-effort CPU-cluster multiprocessing to a multicellularity evolution model, which provides 92\% scaling efficiency at 64 processes ($2.1\times$ speedup) and exhibits robust median quality of service, even under hardware anomalies.
The second case study examines WSE-based simulations, demonstrating best-effort strategies to track spatiotemporal population history --- through sparse, asynchronous device-to-host sampling that tolerates hardware faults.
In sum, across potential forms and scopes of best-effort relaxation, we argue that digital evolution is uniquely positioned to contribute in developing post-deterministic HPC paradigms.
}
 
\section{Introduction}
\label{moreno:sec:introduction}
\vspace{-2.0ex}

Relative to instances of digital evolution\MyIndex{digital evolution} \textit{in silico}, biological evolution operates at nearly incomprehensible scale.
It has been estimated that the biosphere hosts $5 \times 10^{30}$ living cells and on the order of $5.3 \times 10^{37}$ DNA base pairs --- a workload analogized to the storage and processing capacity of between $10^{20}$ and $10^{21}$ supercomputers circa 2015 \cite{Landenmark2015}.
As such, many interesting biological processes remain out of reach of satisfactory modeling experiments.
Rapid generational turnover in digital evolution often comes at the cost of small population sizes \cite{moreno2022engineering}.
This limitation hinders computational experiments incorporating many-species eco-evolutionary dynamics and evolutionary transitions in individuality (e.g., multicellularity, eusociality) \cite{penczykowski2016understanding,alizon2012modelling,schreiber2021cross}.
In the field of artificial life\MyIndex{artificial life}, such cross-scale processes are of great interest in exploring how evolution can produce ongoing complexity, novelty, and adaptation \cite{taylor2016open,dolson2021digital,taylor2019evolutionary,ackley2014indefinitely}.
Indeed, evidence suggests such experiments have been meaningfully constrained by computational scale \cite{channon2019maximum}.

Analogous limitations in computational scale, in retrospect, turn out to have profoundly influenced connectionist approaches to artificial intelligence \cite{kaplan2020scaling}.
Deep learning has gained prominence in tandem with growth in training set sizes, parameter counts, and training FLOPs \cite{marcus2018deep}.
Since then, AI/ML workloads have brought to market powerful next-generation compute accelerators\MyIndex{hardware accelerators} affording up to hundreds of thousands of processing cores (e.g., by Tenstorrent, Groq, Graphcore, SambaNova, and Cerebras) \cite{zhang2016cambricon,emani2021accelerating,jia2019dissecting,medina2020habana}.
These accelerator devices are anticipated to drive advances in both agent-based modeling (ABM) and high-performance computing (HPC)\MyIndex{high-performance computing} writ large \cite{perumalla2022computer,VanEssendelft2025}, and opportunity exists to benefit digital evolution.

To maximize compute throughput, emerging accelerator platforms exploit highly distributed dataflow architectures --- though this can come at the cost of on-device memory scarcity and locality.
For instance, to pack 880,000 processors onto a single chip, the Cerebras Wafer-Scale Engine (WSE)\MyIndex{wafer-scale engine} platform provisions only 48kB memory per processor, introduces multi-hop message routing, and allows data export only from the chip periphery.
These constraints mirror broader trends in HPC where data storage and movement are becoming more expensive relative to compute \cite{Buluc2021,khan2021analysis,gholami2024ai}.

Effectively harnessing AI/ML accelerators for scientific computing thus poses substantial, but broadly pertinent, engineering challenges.
In the case of digital evolution, a key challenge is tracking population dynamics across a vast, highly distributed fabric of memory-constrained processors.
Such data collection can be essential to digital evolution work.
Hypothesis-driven computational experiments are scientifically useful only insofar as they are interpretable and observable \cite{casti1997wouldbe}.
Likewise, in the context of application-oriented evolutionary computation (EC)\MyIndex{evolutionary computation}, diagnostic information is necessary to troubleshoot, tune parameters, and develop theory \cite{hernandez2022dossier,Hooker1995}.

In some regards, computational costs and constraints arising around data storage and movement resemble the logistical costs and constraints faced in research studying real-world systems.
Rather than ``complete,'' deterministic data observability, most real-world studies instead rely on dynamic, query-driven sampling --- a paradigm we term ``inferential observability.''
Such approaches can degrade gracefully under resource scarcity and intermittent disruptions.

By contrast, most computational research assumes complete observability of model state.
Indeed, the ability to measure the otherwise infeasible is a major benefit of computational experiments \cite{lenski2003evolutionary}.
However, past a certain point, data precision has essentially negligible practical value, such that trading some precision for efficient scaling and hardware accelerator compatibility might be highly worthwhile.

\begin{figure}
\captionsetup[sub]{skip=2pt}

\centering
\begin{subfigure}{\linewidth}
  \includegraphics[width=\linewidth,trim={0 0.2cm 0.2cm 0}]{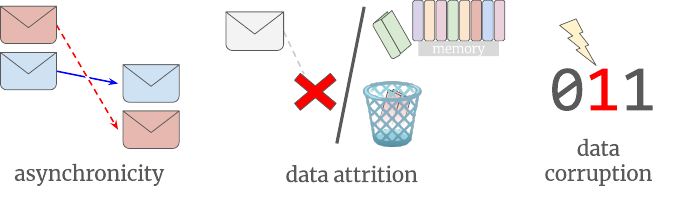}
  \caption{types of best-effort relaxation}
  \label{moreno:fig:best-effort:types}
\end{subfigure}

\vspace{2ex}

\begin{subfigure}{\linewidth}
  \includegraphics[width=\linewidth]{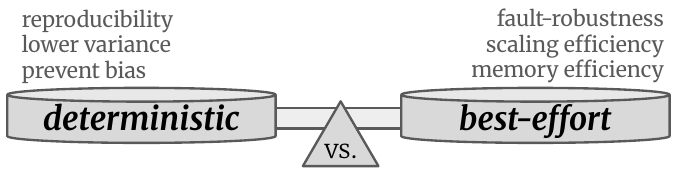}
    \caption{practical trade-offs of best-effort relaxations}
    \label{moreno:fig:best-effort:tradeoffs}
\end{subfigure}

\vspace{2ex}

\begin{subfigure}{\linewidth}
  \includegraphics[width=\linewidth]{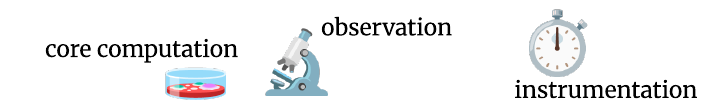}
    \caption{conceptual components of a best-effort system}
    \label{moreno:fig:best-effort:components}
\end{subfigure}

\vspace{2ex}

\begin{subfigure}{\linewidth}
  \includegraphics[width=\linewidth]{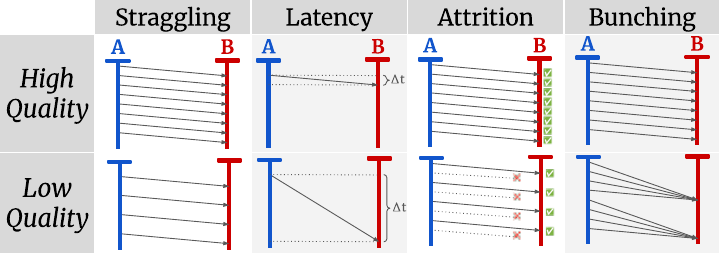}
    \caption{dimensions of quality of service (QoS) in best-effort computation}
    \label{moreno:fig:best-effort:qos}
\end{subfigure}

\vspace{0.5ex}

\caption{
\textbf{Conceptual framework for best-effort computing.}
\footnotesize
Best-effort computing strategies may tolerate asynchronicity, data attrition (e.g., packet loss, dynamic eviction, crashout), or data corruption (panel \subref{moreno:fig:best-effort:types}).
While such strategies can improve robustness and efficiency, departing from a reliable deterministic execution model sacrifices computational reproducibility and can degrade computational results by introducing noise and bias (panel \subref{moreno:fig:best-effort:tradeoffs}).
In the context of scientific computing, a distinction can be drawn between an \textit{underlying computation} (e.g., a simulation kernel) and \textit{observations} performed to document its behavior (panel \subref{moreno:fig:best-effort:components});
to balance efficiency and integrity, best-effort relaxation may be restricted to just data collection (``inferential observability'').
As a best practice, projects incorporating best-effort strategies should consider \textit{instrumentation} to monitor runtime behavior of relaxations (panel \subref{moreno:fig:best-effort:components}).
Panel \subref{moreno:fig:best-effort:qos} suggests several quality of service metrics that may be collected.
}
\label{moreno:fig:best-effort}
\end{figure}
 
Here, we explore incorporating \textit{best-effort computing}\MyIndex{best-effort computing} approaches into digital evolution work.
This paradigm broadens processing beyond deterministic logical transforms --- a near-universal assumption of scientific computing \cite{chakradhar2010best,rinard2012unsynchronized}, except in unusual circumstances \cite{desell2009robust}.
Though outside our primary focus, best-effort computing intertwines with related topics of \textit{approximate computing}\MyIndex{approximate computing} (which emphasizes bypassing non-essential computation) and \textit{randomized algorithms}\MyIndex{randomized algorithms} (which typically harness statistical properties of pseudorandomness) \cite{Buluc2021,mittal2016survey,menon2023approximate}.

Best-effort relaxations can allow \textit{asynchronicity} (e.g., transmission delays, data races, jitter, straggler effects), \textit{data attrition} (e.g., memory constraint, transmission failure, hardware crashout), and \textit{data corruption} (e.g., soft errors, partial writes, analog approximation) (Figure \ref{moreno:fig:best-effort:types}).
Due to the economy of error detection in digital computing \cite{hamming1950error} (e.g., via parity, checksums, duplicate operations, inverse checks, or skeptical assertions), we narrow focus to handle data corruption as simple data attrition \cite{ackley2013beyond}.

In existing work, best-effort relaxations have been shown to improve speed \cite{kasap2018dynamic,chakrapani2008probabilistic}, energy efficiency \cite{chakrapani2008probabilistic,bocquet2018memory}, solution quality \cite{rinard2013parallel}, and scalability \cite{meng2009best}.  Best-effort approaches, however, involve important drawbacks (Figure \ref{moreno:fig:best-effort:tradeoffs}).
Most obviously, best-effort computation abandons bitwise reproducibility\MyIndex{reproducibility} --- the guarantee that repeating a computation will produce an identical result \cite{menon2025reproducibility}.
In addition to introducing stochastic noise, such relaxation can undermine qualitative integrity and reproducibility by unpredictably biasing results.
For instance, in the context of EC, asynchronous fitness evaluation\MyIndex{asynchronous evaluation} can implicitly disfavor slower-to-evaluate solutions \cite{scott2015understanding,scott2016evaluation,scott2022avoiding,guijt2023impact,karns2025evaluation}.
In tandem, algorithms become more opaque, as they are no longer fully described independent from an underlying execution context.

Given these drawbacks, prudent integration of best-effort approaches will require careful consideration.
We suggest, in particular, two mitigations: domain-appropriate scoping of allowed relaxations and, for relaxations that are incorporated, monitoring of runtime quality of service (QoS)\MyIndex{quality of service} \cite{karakus2017quality}.

To structure discussion of these mitigations, we frame a best-effort system's runtime activity in three parts: (1) an underlying \textit{core computation}, (2) \textit{observation} of that computation's intermediate and final states, and (3) \textit{instrumentation} of runtime behavior (Figure \ref{moreno:fig:best-effort:components}).
In the context of digital evolution, \textit{core computation} would comprise runtime operations that influence an end-state population and \textit{observation} would include operations to record lineage histories, discrete evolutionary events, or aggregate population statistics.
As for \textit{instrumentation}, possibilities include send/receive counters, timestamps, profiling traces, event logs, and records of what hardware is used.

In scoping allowed relaxations, distinction can be drawn between \textit{core computation} and \textit{observation}.
For core computation, early irregularities propagate through later steps, which may spiral out of control.
By contrast, although systematic biases may still be introduced, irregularity in observation affects only \textit{post hoc} analysis --- where it may often be more readily reasoned about and accounted for.
Accepting such best-effort observation can, in exchange, allow efficiencies in dynamically exploiting convenient data locality or available openings in unused bandwidth and memory, as well as fault robustness and capped-budget utilization guarantees.
Such an \textit{inferential observability} paradigm thus strikes an intermediate compromise, which may suit a broader swath of use cases.

As for aspects of runtime behavior that should be instrumented, an exhaustive treatment remains for future work.
In this initial exploration, we consider four broad QoS dimensions (Figure \ref{moreno:fig:best-effort:qos}):
\begin{itemize}
  \item \textit{straggling}, increase in time elapsed per computational operation processed;
  \item \textit{latency}, communication delay between sender and receiver;
  \item \textit{attrition}, data discarded or lost --- both in flight and at rest; and
  \item \textit{bunching}, irregularity in the timing of communication.
  \unskip\footnote{Section \ref{moreno:sec:case1:metrics} defines metrics to measure each QoS dimension.
  }
\end{itemize}

Along these dimensions, runtime behavior may be considered in average, variance, extremes, and spatiotemporal distribution (i.e., across physical hardware and logical simulation content).
In practice, QoS measurement can be used to help tune runtime configuration within acceptable bounds, flag anomalous hardware components, screen for problematic biases, aid qualitative reproducibility, and interpret computational results.
We also suggest QoS be considered alongside execution speed in scaling benchmarks, to characterize performance-fidelity tradeoffs (Section \ref{moreno:sec:case1:qos-scaling}).

To explore how best-effort relaxations of HPC behave in practice, we present two case studies.
In the first, we apply QoS metrics to assess performance characteristics of a best-effort multiprocessing framework developed to support digital evolution experiments.
In the second case study, we demonstrate best-effort strategies for approximate tracking of evolutionary history in action on the WSE platform.

\vspace{-4.0ex}
\section{Best-effort Case Study: Cluster-based HPC} \label{moreno:sec:case1}
\vspace{-2.0ex}

\begin{figure}
\captionsetup[sub]{skip=2pt}

\centering
\begin{subfigure}{\linewidth}
  \includegraphics[width=\linewidth]{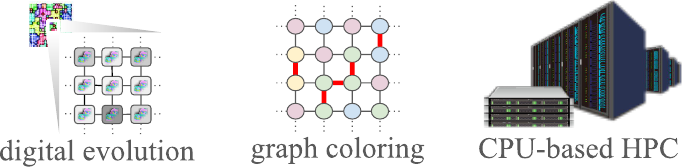}
  \caption{benchmark problem domains and hardware platform}
  \label{moreno:fig:cpu:architecture}
\end{subfigure}

\vspace{0.5ex}
\centering
\begin{subfigure}{\linewidth}
  \includegraphics[width=\linewidth]{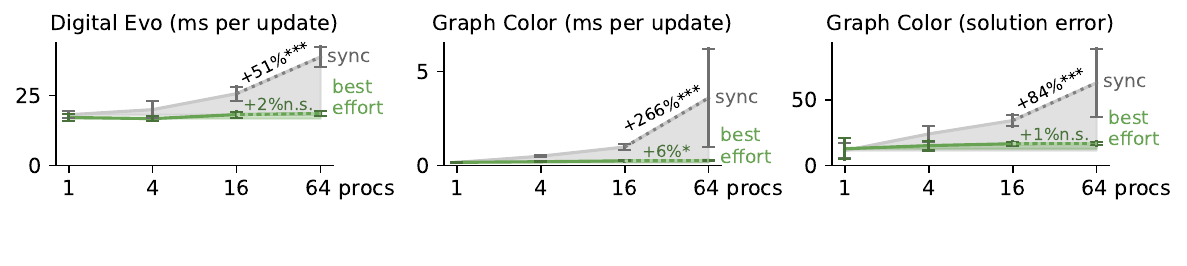}
\vspace{-7ex}
  \caption{comparison of best-effort vs. synchronous performance scaling; higher values worse}
  \label{moreno:fig:cpu:performance}
\end{subfigure}

\vspace{0.5ex}

\begin{subfigure}{\linewidth}
  \centering
\includegraphics[width=\linewidth]{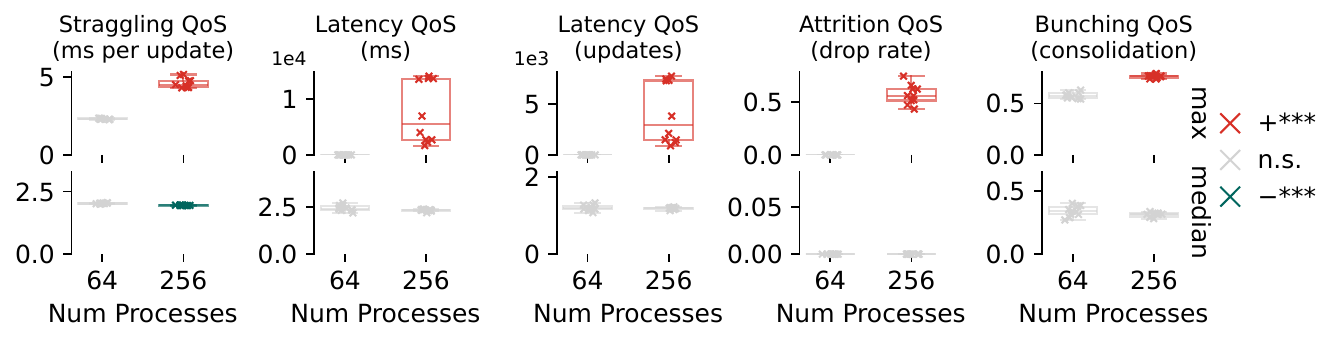}
  \vspace{-3.7ex}
  \caption{stability of median and extreme QoS under best-effort scaling; higher values worse}
  \label{moreno:fig:cpu:qos-scaling}
\end{subfigure}

\vspace{0.5ex}

\begin{subfigure}{\linewidth}
  \centering
  \includegraphics[width=\linewidth]{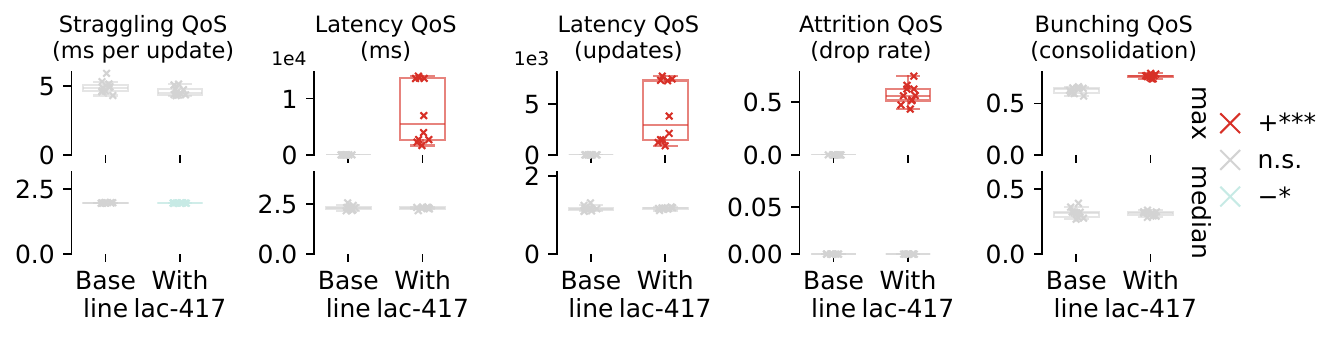}
\vspace{-3.7ex}
  \caption{effect of anomalous node on median and extreme QoS; higher values worse}
  \label{moreno:fig:cpu:lac417}
\end{subfigure}

\vspace{0.5ex}

\begin{subfigure}[b]{0.45\linewidth}
\begin{minipage}{\linewidth}
\centering
\begin{minipage}[t]{0.65\linewidth}
\vspace{0pt} \includegraphics[width=1.1in,angle=90]{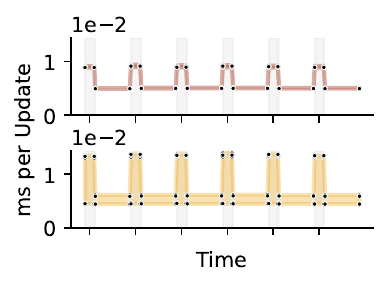}\end{minipage}\begin{minipage}[t]{0.28\linewidth}
  \vspace{0pt} \centering
  \fontsize{6.5}{6.5}\selectfont
  ~\\[2.12em]
  NUMA-\\[0.39em]
  symmetric\\[1.98em]
  NUMA-\\[0.39em]
  asymmetric
\end{minipage}
\vspace{-1.5ex}
\end{minipage}
\caption{intranode snapshot cycle profiles}
\label{moreno:fig:cpu:intranode}
\end{subfigure}\begin{subfigure}[b]{0.3\linewidth}
\centering
  \includegraphics[height=1in]{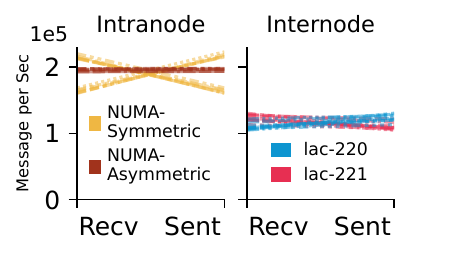}\vspace{-1.2ex}
  \caption{msg. balance diagnostic}
  \label{moreno:fig:cpu:balance}
\end{subfigure}\begin{subfigure}[b]{0.25\linewidth}
\centering
  \includegraphics[height=1in]{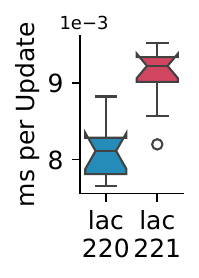}\vspace{-1.2ex}
  \caption{internode perf.}
  \label{moreno:fig:cpu:internode}
\end{subfigure}

\vspace{-0.5ex}

\caption{
\textbf{Cluster-based HPC Case Study.}
\footnotesize
Case study examines digital evolution and graph coloring benchmarks on a CPU-based compute cluster (panel \subref{moreno:fig:cpu:architecture}).
Compared to synchronous evaluation, best-effort relaxations improve scaling efficiency (panel \subref{moreno:fig:cpu:performance} one process per node; error bars SD).
For graph coloring, median QoS remained stable between allocation sizes but magnitude of worst-case QoS disruptions grew (panel \subref{moreno:fig:cpu:qos-scaling}).
In further 256-process experiments testing impact of anomalous hardware, median QoS remained stable with and without low-QoS node \texttt{lac-417} (panel \subref{moreno:fig:cpu:lac417}).
In a final experiment, QoS metrics were collected between graph coloring processes resident on the same node (``intranode'') or split between two nodes (``internode'').
In both cases, negative correlation arose between message send and receive rates (panel \subref{moreno:fig:cpu:balance}).
For within-node configuration, this imbalance likely reflects asymmetric Non-Uniform Memory Access (NUMA) disruptions.
Filled strips in panel \subref{moreno:fig:cpu:intranode} show spread between fastest and slowest process (higher values worse; replicates overlaid), which respond to QoS polling thread activity (shaded intervals).
For internode configuration, send/receive imbalance relates to disparate process speeds between host nodes (panel \subref{moreno:fig:cpu:internode}).
Significance annotations: * $p<0.05$ and *** $p < 0.001$.
}
\label{moreno:fig:cpu}
\end{figure}
 
Evolutionary transitions in individuality, such as multicellularity and eusociality, represent a key facet of biological evolution's surprising constructive power \cite{taylor2016open}.
Digital evolution experiments exploring how these transitions unfold can be computationally demanding: life history of each multicellular collective may encompass an independent subpopulation of replicators --- possibly themselves engaged in interaction-intensive developmental and physiological processes.

In this case study, we test the performance characteristics of a best-effort multiprocessing framework developed to support the DISHTINY platform, which targets such experiments \cite{moreno2019toward}.
This system tracks a fixed-capacity population of virtual cells controlled by event-driven linear genetic programs (GP)\MyIndex{genetic programming}\MyIndex{linear genetic programming} \cite{lalejini2018evolving}, which may form cooperative groups with local neighbors \cite{moreno2021exploring}.

To support multiprocessing, DISHTINY routes all cell-cell interactions through a communication layer implemented by the Conduit library \cite{moreno2021conduit}.
\unskip\footnote{The Conduit library is one among many frameworks that have arisen to offer useful parallel and distributed programming abstractions, including task-based frameworks \cite{kale1993charm++,bauer2012legion,blumofe1996cilk,reinders2007intel} or domain-specific programming languages \cite{el2006upc,chamberlain2007parallel}.
However, these frameworks generally assume a deterministic relationship between tasks or otherwise forbid data races.
}
This multiprocessing framework exposes a best-effort interface backed by MPI intrinsics \cite{gropp1996high}.
When two neighboring cells reside on different processes, a configurable-capacity send buffer asynchronously dispatches triggered interactions and synchronizes local state.
When available send buffer capacity fills, communication may be configured to drop rather than block.
These best-effort relaxations allow per-processor subpopulations to proceed in a fully asynchronous manner.
\unskip\footnote{MPI guarantees delivery of dispatched messages --- network-layer best-effort relaxations would involve lower-level protocols like InfiniBand Unreliable Datagrams \cite{kashyap2006ip,koop2007high}.
}

Our case study examines this best-effort framework, testing four questions: (1) can best-effort relaxations improve scaling efficiency, (2) can quality of service metrics detect runtime irregularities arising from relaxations, (3) does typical quality of service (QoS) scale robustly, and (4) is typical quality of service robust to hardware anomalies.

\vspace{-4.0ex}
\subsection{Materials and Methods}
\label{moreno:sec:case1:design}
\vspace{-2.0ex}

Default-condition DISHTINY simulations were used as an example digital evolution workload for runtime profiling of best-effort relaxations.
Underlying Conduit communication was instrumented with counters to tally sends, receives, and round-trip touches
\unskip\footnote{To avoid clock synchronization issues, no cross-process duration timings were taken.
Instead, runtime instrumentation maintained independent zero-initialized ``touch counters,'' appended to all dispatched messages.
By incrementing this value upon receipt, it increases by two for each successful round trip completed.
}.
In addition to recording final counter values at completion, a background thread polled counters once per minute to collect values bookending a one-second ``snapshot'' window.

To profile compute-heavy conditions representative of island-model EC\MyIndex{island model}, DISHTINY profiling assigned a grid size of 3{\small,}600 cells per process (a widely used default \cite{ofria2004avida}).
For efficiency, Conduit was configured to consolidate communication between processes to single bulk transfers once per update cycle.

To test communication-intensive workloads, we supplemented DISHTINY trials with a lightweight graph coloring solver \cite{leith2012wlan}.
Nodes, arranged in a toroidal lattice, begin with one of three randomly assigned colors; a problem size of 2{\small,}048 nodes per process was used.
Under this solver, nodes conflicting with neighbors probabilistically self-assign a new color.
As another benefit, the number of remaining same-color conflicts provides an explicit measure of solution quality.

Benchmark and profiling trials were replicated across 10 independent executions within a given SLURM allocation.
Within each replicate, performance metrics are reported as mean execution speed across processes; for QoS, median and maximum are reported.
QoS experiments utilized Michigan State University's \texttt{lac} cluster, consisting of dual-socket nodes with 14-core Intel Xeon E5-2680 v4 CPUs @ 2.40 GHz (28 cores total) --- each socket acting as a distinct Non-Uniform Memory Access domain --- joined by InfiniBand interconnect.
Hostnames and node configurations are recorded for each collected observation.
Further details on methods are included with Supplemental Material \cite{moreno_2026_trust}.

\vspace{-4.0ex}
\subsection{Quality of Service (QoS) Metrics}
\label{moreno:sec:case1:metrics}
\vspace{-2.0ex}

Reported metrics measure four dimensions of runtime QoS (Figure \ref{moreno:fig:best-effort:qos}):
\begin{enumerate}
\item \textbf{Straggling QoS}: \textit{update time}, amount of walltime elapsed per update cycle;
\item \textbf{Latency QoS}: \textit{update-latency} or \textit{walltime-latency}, duration elapsed between message dispatch and delivery, estimated via a round-trip counter;
\item \textbf{Attrition QoS}: \textit{drop rate}, proportion of sent messages failing to be delivered; and
\item \textbf{Bunching QoS}: \textit{consolidation}, the proportion of possible independent delivery events eliminated by batching.
\end{enumerate}

To ensure consistent interpretation, all metrics quantify QoS degradation --- such that higher values indicate worse quality of service.
To reflect runtime volatility, QoS measurements were captured during 1-second sample windows.
Formulas to calculate these metrics from timestamped instrumentation counter values are provided in Supplemental Material \cite{moreno_2026_trust}.

\vspace{-4.0ex}
\subsection{Best-effort Scaling Efficiency}
\label{moreno:sec:case1:scaling-efficiency}
\vspace{-2.0ex}

To test whether best-effort relaxations can improve scaling efficiency\MyIndex{scalability}, we compared execution speeds across trials harnessing 1, 4, 16, and 64 processes.
Allocations were configured with one process per cluster node at constant problem size per process (i.e., weak scaling).

Best-effort relaxation substantially improved scaling efficiency of the compute-intensive digital evolution workload.
Scaling DISHTINY between 1 and 64 processes, best-effort relaxation maintained 92\% efficiency --- significantly better than 47\% efficiency under synchronous execution (Mann-Whitney test, $p<0.001$; leftmost panel, Figure \ref{moreno:fig:cpu:performance}).
Past 16 processes, best-effort relaxation maintained stable performance, while synchronous execution degraded by 34\% (Mann-Whitney tests, $\alpha = 0.05$ and $p < 0.001$; leftmost panel of Figure \ref{moreno:fig:cpu:performance}).
For this workload, benefit from asynchrony likely stems from easing imbalances in evaluation of heterogeneous GP content.

Communication-intensive graph-coloring workloads also benefited from best-effort relaxation.
Between 1 and 64 processes, execution throughput degraded 95\% under synchronous execution --- while best-effort evaluation proved more efficient, degrading only 36\% (Mann-Whitney test, $p < 0.001$; center panel, Figure \ref{moreno:fig:cpu:performance}).
Likewise, solution error fared better under best-effort evaluation ($1.3\times$ vs. $4.5\times$ increase; right panel, Figure \ref{moreno:fig:cpu:performance}).
Scaling past 16 processes, solution error remained stable under best-effort relaxation (Mann-Whitney test, $\alpha = 0.05$), but degraded $0.8\times$ under synchronous evaluation (Mann-Whitney test, $p < 0.001$).
In this window, execution speed degraded 6\% under best-effort relaxation (Mann-Whitney test, $p < 0.05$) but 72\% under synchronous evaluation (Mann-Whitney test, $p < 0.001$).

Across the board, best-effort relaxation improved efficiency at 64-process scale: providing a $2.1\times$ speedup to compute-intensive workload (DISHTINY) and --- for communication-intensive workload (graph coloring) --- $12.5\times$ speedup with 73\% reduction in solution error.
In all three cases, improvement is significant (Mann-Whitney tests, all $p < 0.001$).

\vspace{-4.0ex}
\subsection{Quality of Service Scaling}
\label{moreno:sec:case1:qos-scaling}
\vspace{-2.0ex}

Having seen best-effort relaxation benefit scaling efficiency, we next set out to test whether efficiency came at the cost of degraded QoS.
For these experiments, we compared 64- and 256-process allocations, the largest available from our resource provider.
\unskip\footnote{On account of original plans to incorporate threading-based trials, preceding experiments in Section \ref{moreno:sec:case1:scaling-efficiency} were run only to 64 processes.
}
Graph coloring benchmarks were conducted to exercise communication intensity.

Figure \ref{moreno:fig:cpu:qos-scaling} compares maximum (i.e., worst within allocation) and median (i.e., typical within allocation) QoS metrics between 64- and 256-process jobs.
All reported QoS metrics --- unsurprisingly --- exhibited more extreme tail values in larger process pools (Mann-Whitney tests, all $p < 0.001$; Figure \ref{moreno:fig:cpu:qos-scaling}, top row).
By contrast, median QoS tolerates scaling between 64 and 256 processes (Figure \ref{moreno:fig:cpu:qos-scaling}, bottom row).
Across metrics, no detectable degradation arose between process pool sizes (Mann-Whitney tests, $\alpha = 0.05$), and straggling, in fact, reduced by 4\% (Mann-Whitney test, $p < 0.001$).

Scaling between 64- and 256-process allocations under even more challenging conditions with extreme communication intensity (1 graph coloring node per process) produced similar QoS outcomes.
Among median QoS readings, only straggling detectably worsened by 8\% (Mann-Whitney tests, $\alpha = 0.05$; Supplemental Figure 1 \cite{moreno_2026_trust}).

\vspace{-4.0ex}
\subsection{Robustness to Anomalous Hardware}
\label{moreno:sec:with-lac-417-vs-sans-lac-417}
\vspace{-2.0ex}

The extreme magnitude of QoS outliers observed in 256-process allocations prompted closer inspection, which revealed that extreme outliers all involved cluster node \texttt{lac-417}.
So, we performed further tests acquiring two separate 256-process allocations on the lac cluster: one including \texttt{lac-417} and one excluding \texttt{lac-417}.

Figure \ref{moreno:fig:cpu:lac417} compares the distributions of QoS metrics between 256-process allocations with and without \texttt{lac-417}.
For straggling QoS, worst-case slowdowns were comparable between allocations (Mann-Whitney test, $\alpha = 0.05$).
For other measures, \texttt{lac-417} introduced up to orders-of-magnitude more extreme outlying QoS values (Mann-Whitney tests, all $p<0.001$; Figure \ref{moreno:fig:cpu:lac417}).

Despite extreme outliers, median QoS values remained stable across all metrics (Mann-Whitney tests, $\alpha = 0.05$; Figure \ref{moreno:fig:cpu:lac417}, lower panel).
Incorporating \texttt{lac-417}, in fact, produced a marginal 0.5\% reduction in straggling (Mann-Whitney test, $p=0.04$).
Thus, in this case, global QoS robustly tolerated local disruptions.

\vspace{-4.0ex}
\subsection{Diagnosis of QoS Asymmetry}
\label{moreno:sec:case1:asymmetry}
\vspace{-2.0ex}

Having examined QoS metrics in aggregate across process allocations, we next sought to examine whether these metrics could detect and diagnose localized QoS asymmetries emerging at the level of individual processes.
In a final pair of experiments, we benchmarked two-process allocations --- either split between cluster nodes (``internode'') or co-located on the same node (``intranode'').
To maximize communication intensity, we tested execution with one graph coloring node per process.

As a screen for producer-consumer imbalance, where one process falls behind and becomes livelocked clearing incoming messages from the other, we examined whether processes receiving more messages tended to send fewer outgoing messages.

Under both inter- and intranode configurations, we indeed found significant anticorrelation between incoming and outgoing traffic (Figure \ref{moreno:fig:cpu:balance}; GLM, $p < 0.001$).
Interestingly, in the case of intranode communication, an apparent bimodal distribution arose, where traffic was either balanced or strongly imbalanced (left facet, Figure \ref{moreno:fig:cpu:balance}).

In the between-node experiment, imbalance arose from an underlying 12\% disparity in execution speed between host nodes (Figure \ref{moreno:fig:cpu:internode}); after accounting for this factor, no detectable traffic imbalance remained (GLM, $p = 0.15$).

In the within-node case, traffic imbalance appears likely to have arisen as an artifact of Non-Uniform Memory Access (NUMA) architecture.
Because host cluster nodes use two independent CPU sockets, conflicting memory accesses can trigger cache invalidations that stall execution.
Figure \ref{moreno:fig:cpu:intranode} visualizes straggling QoS within process pairs over the 5-minute course of execution trials.
Across all replicates, at least one process slows during one-second snapshot windows --- consistent with a scenario where the background polling thread violates NUMA.
Where incoming/outgoing imbalance arose, only one process slows; otherwise, both processes slow.
Thus, observed imbalance appears likely due to asymmetric NUMA effects.

The first step to resolving local asymmetries and anomalies is detecting and diagnosing them.
In the case of NUMA-driven imbalance, allocation flags can be used to assign process residency.
In the case of the rate-driven imbalance, outgoing message rates could be throttled to countervailing traffic.
Alternatively, such asymmetries could simply be recorded and factored into interpretation of computational results.
 
\vspace{-4.0ex}
\section{Best-effort Case Study: Accelerator-based HPC}
\label{moreno:sec:case2}
\vspace{-2.0ex}

\begin{figure}
\captionsetup[sub]{skip=2pt}

\centering
\begin{subfigure}{\linewidth}
  \includegraphics[width=\linewidth]{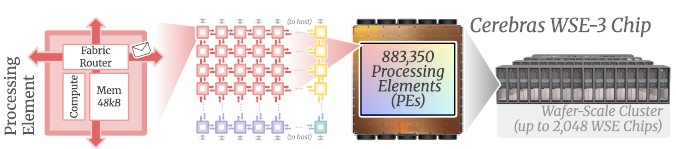}
  \caption{Wafer-Scale Engine architecture}
  \label{moreno:fig:wse:architecture}
\end{subfigure}

\vspace{2ex}

\begin{minipage}{0.3\linewidth}
\begin{subfigure}{\linewidth}
  \includegraphics[width=\linewidth]{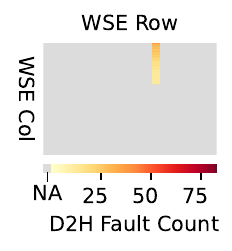}
  \caption{encountered h/w anomaly}
  \label{moreno:fig:wse:fault}
\end{subfigure}\end{minipage}\begin{minipage}{0.7\linewidth}
\begin{subfigure}{\linewidth}
  \includegraphics[width=\linewidth]{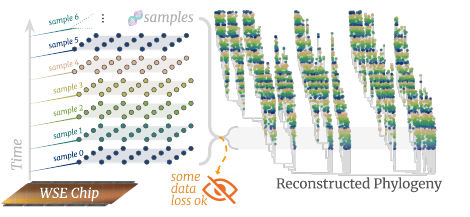}
  \caption{sampling-based estimation of evolutionary history}
  \label{moreno:fig:wse:phylogeny}
\end{subfigure}
\end{minipage}

\vspace{2ex}

\begin{subfigure}{\linewidth}
  \includegraphics[width=\linewidth]{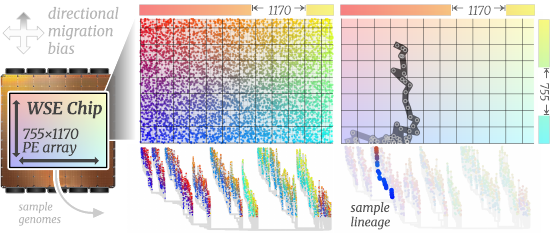}
    \caption{directional migration bias along lineage histories under neutral evolution}
    \label{moreno:fig:wse:bias}
\end{subfigure}

\vspace{2ex}

\begin{subfigure}{\linewidth}
  \includegraphics[width=\linewidth]{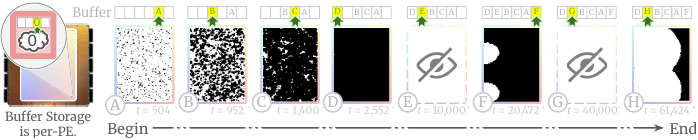}
    \caption{fixed-capacity strategy for best-effort recording of spatiotemporal hypermutator dynamics}
    \label{moreno:fig:wse:timeseries}
\end{subfigure}

\vspace{0.5ex}

\caption{
\textbf{Accelerator-based HPC Case Study.}
\footnotesize
Wafer-Scale Engine (WSE) architecture arranges independent processing elements (PEs) in a grid lattice (panel \subref{moreno:fig:wse:architecture}).
To accommodate constraints in host-device communication and on-device memory, lineage histories can be estimated using a best-effort approach via barcode-like markers in sparsely sampled genomes (panel \subref{moreno:fig:wse:phylogeny}).
Such a strategy gracefully tolerates missing data, such as occurred in a persistent device-to-host transfer fault on one WSE-3 chip (panel \subref{moreno:fig:wse:fault}).
Migration between per-PE subpopulations is also conducted asynchronously;
before correction, a small directional bias in migration rates biased long-term lineage histories (panel \subref{moreno:fig:wse:bias}).
In experiments exploring the evolution of hypermutator traits, where recording duration was unknown \textit{a priori}, time series records of local hypermutator prevalence were curated on a best-effort basis using a generalized ring buffer (panel \subref{moreno:fig:wse:timeseries}, hypermutators in black; video at \url{https://hopth.us/ej}).
}
\label{moreno:fig:wse}
\end{figure}
 
In addition to synchronization inefficiencies, best-effort strategies can also help overcome on-device memory scarcity and data loss from hardware faults, two factors that have come to the fore in ongoing work with agent-based evolution simulations on the Cerebras Wafer-Scale Engine (WSE) platform.

The third generation WSE comprises a $755 \times 1{\small,}170$ lattice of 883{\small,}350 networked compute cores (Processing Elements or PEs; Figure \ref{moreno:fig:wse:architecture}).
PEs run independently with CPU-like control flow, and use on-chip routing to communicate between lattice neighbors. On-device memory, 44GB in aggregate, is sharded across PEs --- providing each a private memory bank.
Thus, work on WSE must contend with scarce memory (48kB per PE), extreme decentralization (direct communication only with local neighbors), and restricted host-device connectivity (only at lattice periphery).
Physical stresses from kilowatts of thermal loading \cite{la2020cerebras}, coupled with sheer component count of transistors, also make hardware faults a marked concern.

This case study explores incorporation of best-effort strategies in wafer-scale digital evolution experiments, drawing from two ongoing projects:
\begin{enumerate}
\item work with a dummy agent model, as a vehicle to test methodology for tracking evolutionary history \cite{singhvi2025scalable}, and
\item work scaling up an existing model of mutator allele evolution to explore dynamics in large, spatially distributed populations \cite{raynes2018sign,moreno2025hypermutatorsc}.
\end{enumerate}
Both of these applications target WSE using an island model scheme \cite{bennett1999building}, where each PE hosts an independent subpopulation and exchanges migrants with neighbors.
Further detail on configurations appears in supplementary material \cite{moreno_2026_trust}.
Case study material here is presented in four parts,
\begin{enumerate}
\item overview of approach for approximate tracking of evolutionary history (Section \ref{moreno:sec:case2:hstrat});
\item application of this approach to tolerate sparse, asynchronous device-to-host data collection and a persistent localized data-corruption fault (Section \ref{moreno:sec:case2:missing});
\item use of spatiotemporal evolutionary tracking to identify emergent biases in population dynamics (Section \ref{moreno:sec:case2:bias}); and
\item application of a generalized ring buffer approach to efficiently record unknown-duration time-series data within a fixed memory budget (Section \ref{moreno:sec:case2:timeseries}).
\end{enumerate}

\vspace{-4.0ex}
\subsection{Reconstruction-based Estimation of Evolutionary History}
\label{moreno:sec:case2:hstrat}
\vspace{-2.0ex}

In WSE-based experiments, agent migrations across the chip pose substantial logistical challenges in tracking evolutionary history as it unfolds \cite{moreno2024analysis}.
On WSE, instrumentation must be lightweight because memory used comes at the direct cost of simulation capacity.
Even were sufficient memory and bandwidth available, a complete evolutionary history --- upwards of a quadrillion replication events in some simulations \cite{singhvi2025scalable} --- would be impractical in its entirety; instead, means are needed to extract a representative summary.

In evolutionary studies, one key aspect of history is phylogeny\MyIndex{phylogeny} --- the structure of ancestry relationships among organisms \cite{faithConservationEvaluationPhylogenetic1992, STAMATAKIS2005phylogenetics,frenchHostPhylogenyShapes2023,kim2006discovery,lenski2003evolutionary}.
In addition to tracing the history of evolutionary events \cite{dawkins2016ancestor,hedges2015tree,hinchliff2015synthesis}, phylogenetic analysis can test more general questions about the underlying mode and tempo of evolution \cite{felsenstein1985phylogenies}.
Classically, these analyses have been applied to species-level macroevolution; however, population- and organism-level dynamics can also be inferred \cite{genthon2023cell, levy2015quantitative,stadler2013recovering,Moreno2025ecology,lewinsohnStatedependentEvolutionaryModels2023a,nozoe2017inferring}.
Phylogeny data can also be leveraged in proactive efforts to influence the trajectory of an evolving population \cite{Scott2018}.
In the case of application-oriented EC, such strategies can help guide evolution toward desired outcomes \cite{lalejini2024phylogeny,lalejini2024runtime,murphy2008simple,burke2003increased,gabor2018inheritancebased,gabor2018preparing}; phylogeny can also provide informative diagnostics to practitioners \cite{hernandez2022can,shahbandegan2022untangling}.
Finally, digital phylogenies additionally serve as a useful testbed for bioinformatics because of capability to configure ground truth conditions directly \cite{daudey2024aevol,haller2023slim,moreno2025extending}.

To collect ancestry data from WSE experiments, we have adopted strategies inspired by real-world experiments that assume data collection is expensive, limited, and non-deterministic.
Rather than exhaustively tracking parent-child relationships \cite{dolson2023phylotrackpy}, we adopt an approach inspired by bioengineered lineage tracking\MyIndex{lineage tracking} \cite{moreno2022hereditary} --- in which progressive DNA barcode inserts identify closely related organisms by shared commonalities \cite{Masuyama2019,konno2022deep,jones2020inference,kebschull2018cellular,pan2026integrative,jones2024spatiotemporal}.
We implement this approach using single-bit markers as barcodes \cite{moreno2025testing}, stored within a fixed-capacity 64-bit genome region; once full, markers overwrite each other to maintain an approximate record \cite{moreno2024algorithms}.
In contrast to scenarios where the chronological order of random mutations is unknown \cite{gabor2017genealogical,moreno2021case},
the known timing of injected barcodes allows reconstruction of phylogenetic trees by simple agglomerative trie-building over shared prefix sequences \cite{moreno2023toward,singhvi2025scalable}.
While this methodology primarily targets tracking of asexual lineages (i.e., without recombination/crossover), possibilities exist for extensions to sexual phylogenies  \cite{moreno2024methods} which exhibit important structural differences \cite{mcphee2016using,godindubois2024apoget,burlacu2015methods}.

\vspace{-4.0ex}
\subsection{Robustness of Best-effort Estimation of Evolutionary History}
\vspace{-2.0ex}
\label{moreno:sec:case2:missing}

A key advantage of reconstruction-based evolutionary analysis is flexibility with respect to the number of taxa sampled (i.e., phylogeny tree tips).
Evolutionary divergence between a single pair of taxa can be estimated just the same as for much larger collections.
Ultimately, necessary data collection depends on experimental objectives --- for instance, taxa may be sampled through time, or only at the final time point.

Because it tolerates sparse, irregular sampling, reconstruction-based analysis is well suited to best-effort strategies.
In reported runs \cite{singhvi2025scalable}, sample genomes were gathered from WSE-based simulation on a rolling basis using asynchronous device-to-host (D2H) transfers (Figure \ref{moreno:fig:wse:phylogeny}).
This approach allowed available bandwidth to be dynamically balanced with existing on-chip traffic.
Experiments evaluated a population of 226.1 million agents for 5 million generations, running about 8{\small,}750 generations per second walltime.
D2H transfers sampled one genome per PE, and completed approximately 15 times per second --- yielding 9{\small,}000 snapshots over the course of a 10-minute runtime.
This volume of harvested genomes (nearly 8 billion in total) exceeded downstream analysis capacity, and so was thinned on a rolling basis to 1.1 billion genomes ($\sim0.0001\%$ of cumulative individuals).

Another benefit afforded by reconstruction-based analysis is resilience to data loss from hardware faults.
Reported runs used Cerebras Cloud system \texttt{rwse1001-\allowbreak cs-\allowbreak sy01}.
After February 26, 2026, this hardware began to exhibit unreliable D2H transfers along PE row 410 between columns 0 and 248 (Figure \ref{moreno:fig:wse:fault}).
Around 5\% of 32-bit words copied from this region had bit-flip errors, corrupting 0.03\% of overall extracted genomes.
Runtime parity checks confirmed that this phenomenon did not compromise on-device simulation, only D2H export.

In this case, best-effort resilience allowed work to continue uninterrupted over the following two months the chip remained in service.
While clear localization in this case made excluding bad data easy, a lightweight checksum over each PE's exported data could readily handle more generalized transfer faults (barring rare checksum collisions).

Hardware faults\MyIndex{fault tolerance} are inescapable in large-scale HPC \cite{gupta2017failures,navaux2023challenges}, reaching upwards of daily occurrence in some systems \cite{kokolis2025revisiting}.
Faults are typically handled by checkpoint rollbacks \cite{gupta2017failures,zhu2025understanding}.
In some flagship workloads, such checkpointing consumes on the order of 100PB of high-bandwidth disk capacity \cite{frontiere2025cosmological}.
In highly integrated platforms, such as the WSE, challenges are compounded by the fact that small failures can compromise much larger hardware units \cite{cerebras2025yield}.
Where computation can be structured to instead gracefully degrade under disruption, costs can be greatly reduced, and possibilities for suitable computational substrates can be greatly widened \cite{markovi2020physics,ackley2011pursue,dreslinski2010near}.

\vspace{-4.0ex}
\subsection{Detection of Bias in Population Dynamics}
\label{moreno:sec:case2:bias}
\vspace{-2.0ex}

To check the plausibility of reconstructed phylogenies, we examined spatial patterns (i.e., on-chip locality) among related taxa \cite{novembre2008genes,grundler2025geographic}.
Under neutral conditions, lineage histories should constitute a random walk in space \cite{lemey2010phylogeography}.
However, to our surprise, parallel lineages appeared to disproportionately follow similar spatial trajectories (left panel, Figure \ref{moreno:fig:wse:bias}).
Closer inspection of sample lineages revealed that many traversed bottom-to-top across the chip (right panel, Figure \ref{moreno:fig:wse:bias}), suggesting a migration imbalance favoring this direction.

This scenario highlights the importance of screening asymmetries in underlying computation.
In addition to domain-specific inspection of simulation dynamics, systematizing generic QoS-based screens represents an important direction for future work.
Likely, these approaches will prove most effective in conjunction.

\vspace{-4.0ex}
\subsection{Best-effort Time Series Recording}
\label{moreno:sec:case2:timeseries}
\vspace{-2.0ex}

Within evolving microbial populations, genes that elevate mutation rate cause frequent harmful mutations but, on the other hand, create rare beneficial mutations --- a fundamental trade-off known to be sensitive to population size \cite{raynes2018sign}.
To explore how selection acts on such ``mutator alleles'' in spatially structured populations, we harnessed WSE-2 hardware to run large-scale agent-based evolution simulations \cite{moreno2025hypermutatorsc}.

A key challenge in this work was balancing limited on-device memory between simulation content and data recording.
To economize memory use in experiments requiring time-series measurements of mutator prevalence, we applied algorithms that generalize ring buffer storage to manage time-series data \cite{moreno2024algorithms,gunther2014algorithm}.
Successive local samples were recorded as single bits within a flat, fixed-size buffer.
Crucially, because recording duration (e.g., until mutator allele extinction) is unknown \textit{a priori}, this approach dynamically downsamples recording density via overwrites once storage capacity is reached --- strictly bounding memory use.

Among other results, time-series data revealed that even in circumstances where mutator alleles did not reliably fix, they could nonetheless transiently dominate population composition --- in some cases, reaching upwards of 99.9\% prevalence (Figure \ref{moreno:fig:wse:timeseries}).

\vspace{-4.0ex}
\section{Conclusion} \label{moreno:sec:conclusion}
\vspace{-2.0ex}

We have presented two case studies exploring how best-effort strategies can help scale digital evolution work to large compute cluster allocations and accommodate resource constraints to harness processing power of next-generation AI/ML hardware accelerator platforms.

Takeaway findings from our first case study on best-effort cluster-based multiprocessing include:
\begin{enumerate}
\item best-effort communication and synchronization can improve scaling efficiency and solution quality for both homogeneous, communication-heavy workloads and heterogeneous, compute-heavy workloads;
\item median quality of service (QoS) metrics remained generally stable under scale-out, including scenarios that introduce degraded hardware; and
\item QoS instrumentation can help flag and diagnose anomalies in runtime behavior.
\end{enumerate}

Our second case study explored incorporation of best-effort strategies into wafer-scale digital evolution experiments.
Takeaway findings include:
\begin{enumerate}
\item best-effort ``inferential observability'' strategies --- inspired by techniques developed to study natural history --- can provide rich visibility into evolutionary history with lean data movement and storage;
\item such best-effort inferential observability strategies can readily tolerate otherwise disruptive data loss from hardware faults; and
\item domain-specific analyses (e.g., spatiotemporal phylogenies) can complement QoS metrics in screening for computational biases.
\end{enumerate}

Beyond the scope of case study computations, suitability of best-effort relaxations will vary widely between HPC problem domains.
Some domains are clear-cut in favor of the reliable digital machine model \cite{heroux2014toward} --- for example, due to regulatory issues \cite{dongarra2014applied}.
Certain problem characteristics, however, can help tolerate nondeterministic computation.
Heuristic optimization algorithms are good candidates, a notable example being stochastic gradient descent \cite{dean2012large,zhao2019elastic,niu2011hogwild,noel2014dogwild,rhodes2020real}.
Likewise, algorithms relying on randomized methods, which already exploit stochasticity, also perform well \cite{chakrapani2008probabilistic,chakradhar2010best}.

Digital evolution exhibits both of these problem characteristics.
Lending further statistical robustness is the fact that replicators readily repopulate \cite{lyu2021improving,lyu2023online} and, by natural selection, purge degeneracy.
On the other hand, maintaining the fidelity of individual outcomes hinges essentially all-or-nothing on bitwise determinism, given evolution's sensitivity to individual events \cite{ferguson2023potentiating}.

In the case of application-oriented EC, like other optimization domains, worst-case outcomes simply waste resources --- rather than producing junk science.
Even in the case of hypothesis-driven work, however, digital evolution enjoys unusual computational malleability.
While typical simulation seeks to describe a concrete ``real-world'' analog, this is not necessarily the case in digital evolution.
The field often studies evolutionary processes in abstracted terms, by instantiating variation, heredity, and replication within artificial substrates (such as self-replicating computer programs) \cite{pennock2007models,bonabeau1994we}.
Studying ``life-as-it-could-be'' offers broad leeway;
while best-effort relaxations can make a system more difficult to interpret, relaxations do not necessarily invalidate such a model's integrity \textit{ipso facto}.
Even in the case of more literalistic models aligned to a concrete real-world analog, moderate disruptions and irregularities may arguably, in fact, more faithfully reflect biological reality \cite{huberman1993evolutionary}.

At the level of experimental design, perfect reproducibility and observability have certainly, in cases, enabled digital evolution experiments to tackle otherwise intractable questions \cite{pontes2020evolutionary,lenski2003evolutionary,grabowski2013case,dolson2020interpreting,fortuna2019coevolutionary,goldsby2014evolutionary,covert2013experiments,zaman2011rapid,bundy2021footprint,dolson2017spatial}.
However, a perfect reliability digital machine model is not strictly necessary for all such work.
In natural systems, many questions are approached observationally (e.g., phylodynamics, metagenomics, biorepositories) or by limited-scale interventions (e.g., mark-recapture, \textit{ex situ} cultivation, knockout analyses).

For these reasons, digital evolution has a unique opportunity to contribute to the development of post-deterministic computing\MyIndex{post-deterministic computing}.
Indeed, such work exists already in artificial life \cite{ray1995proposal,moreno2021exploring} and EC research \cite{izzo2009parallel,harada2022frequency,liang2024asynchronous}.
Particularly ambitious, and outspoken, exploration of the post-deterministic frontier has been led by Dave Ackley \cite{ackley2014indefinitely}.

Ackley's work \cite{Ackley2023,ganapati2009modular}, alongside many existing studies of best-effort computing \cite{chippa2014scalable,ackley2011homeostatic,cho2012ersa,chakrapani2008probabilistic,rhodes2020real}, has targeted bespoke experimental hardware.
With some exceptions \cite{mayr2019spinnaker210million,pehle2022brainscales}, frontier resources overwhelmingly cater to a common denominator (e.g., for the foreseeable future, AI/ML).
Adapting to the constraints of emerging AI/ML accelerator architectures thus represents a key strategic priority for digital evolution --- and HPC writ large.

Such misappropriation of available resources for unintended purposes is familiar territory.
Indeed, best-effort relaxations will undoubtedly be put to the test by the peculiar tendency of evolution to adversarially exploit its evaluation context \cite{Thompson1997,Lehman2020}.
As developments in conventional hardware continue to shift the cost-benefit balance of strict deterministic computing, lessons learned stand to guide practice that is both rigorous and pragmatic.

\vspace{-2.0ex}
\begin{acknowledgement}
\scriptsize
Computational resources were provided by the MSU Institute for Cyber-Enabled Research and from PSC Neocortex via the ByteBoost training program and under NSF ACCESS Innovative Projects Allocation BIO240102 \cite{buitrago2021neocortex,Boerner2023,Brashear2025}.
This project benefited significantly from open-source scientific software \cite{2020SciPy-NMeth,harris2020array,mckinney-proc-scipy-2010,waskom2021seaborn,hunter2007matplotlib,yang2025downstream,moreno2021signalgp,vostinar2024empirical,moreno2022hstrat,moreno2021conduit,moreno2024wse,moreno2026phyloframe} and graphics \cite{togopic2024_supercomputer,togopic2019_rackserver}.
Thank you also to Mathias Jacquelin and Leighton Wilson at Cerebras Systems.
This material is based upon work supported by the Eric and Wendy Schmidt AI in Science Postdoctoral Fellowship, a Schmidt Sciences program.
This research was supported in part by NSF grants DEB-1655715 and DBI-0939454 and is based upon work supported by NSF GRFP DGE-1424871 and NSF CAREER DEB-2540912.
Any opinions, findings, and conclusions or recommendations expressed in this material are those of the author(s) and do not necessarily reflect the views of the National Science Foundation.
This material is based upon work supported by the U.S. Department of Energy, Office of Science, Office of Advanced Scientific Computing Research (ASCR), under Award Number DE-SC0025634.
This report was prepared as an account of work sponsored by an agency of the United States Government.
Neither the United States Government nor any agency thereof, nor any of their employees, makes any warranty, express or implied, or assumes any legal liability or responsibility for the accuracy, completeness, or usefulness of any information, apparatus, product, or process disclosed, or represents that its use would not infringe privately owned rights.
Reference herein to any specific commercial product, process, or service by trade name, trademark, manufacturer, or otherwise does not necessarily constitute or imply its endorsement, recommendation, or favoring by the United States Government or any agency thereof.
The views and opinions of authors expressed herein do not necessarily state or reflect those of the United States Government or any agency thereof.
\textit{\textbf{AI Use Statement.}}
In this work, AI tools (Claude Code and Google Gemini) were used to assemble manuscript boilerplate, assist data visualization, and proofread; agentic contributions are directly supervised and tracked via commit messages.
\end{acknowledgement}

\vspace{-8.0ex}
\begingroup
\let\oldbibliography\thebibliography

\renewcommand{\thebibliography}[1]{\oldbibliography{#1}\fontsize{7.5pt}{8.6pt}\selectfont
}

\renewcommand{\refname}{References \vspace{-2.0ex}}
\bibliographystyle{spmpsci}

\begin{thebibliography}{100}
\providecommand{\url}[1]{{#1}}
\providecommand{\urlprefix}{URL }
\expandafter\ifx\csname urlstyle\endcsname\relax
  \providecommand{\doi}[1]{DOI~\discretionary{}{}{}#1}\else
  \providecommand{\doi}{DOI~\discretionary{}{}{}\begingroup
  \urlstyle{rm}\Url}\fi

\bibitem{ackley2014indefinitely}
Ackley, D., Small, T.: Indefinitely scalable computing= artificial life
  engineering.
\newblock In: ALIFE 14: The Fourteenth International Conference on the
  Synthesis and Simulation of Living Systems, pp. 606--613. MIT Press (2014).
\newblock \doi{10.7551/978-0-262-32621-6-ch098}

\bibitem{ackley2013beyond}
Ackley, D.H.: Beyond efficiency.
\newblock Communications of the ACM \textbf{56}(10), 38--40 (2013).
\newblock \doi{10.1145/2505340}

\bibitem{Ackley2023}
Ackley, D.H.: A robust programmable replicator for an indefinitely scalable
  machine.
\newblock In: The 2023 Conference on Artificial Life (2023).
\newblock \doi{10.1162/isal_a_00701}

\bibitem{ackley2011pursue}
Ackley, D.H., Cannon, D.C.: Pursue robust indefinite scalability.
\newblock In: Proceedings of the 13th USENIX Conference on Hot Topics in
  Operating Systems, p.~8 (2011)

\bibitem{ackley2011homeostatic}
Ackley, D.H., Williams, L.R.: Homeostatic architectures for robust spatial
  computing.
\newblock In: 2011 Fifth IEEE Conference on Self-Adaptive and Self-Organizing
  Systems Workshops (2011).
\newblock \doi{10.1109/sasow.2011.18}

\bibitem{alizon2012modelling}
Alizon, S., Magnus, C.: Modelling the course of an {HIV} infection: Insights
  from ecology and evolution.
\newblock Viruses \textbf{4}(10), 1984--2013 (2012).
\newblock \doi{10.3390/v4101984}

\bibitem{bauer2012legion}
Bauer, M., Treichler, S., Slaughter, E., Aiken, A.: Legion: Expressing locality
  and independence with logical regions.
\newblock In: SC'12: Proceedings of the International Conference on High
  Performance Computing, Networking, Storage and Analysis, pp. 1--11. {IEEE}
  (2012).
\newblock \doi{10.1109/sc.2012.71}

\bibitem{bennett1999building}
Bennett~III, F.H., Koza, J.R., Shipman, J., Stiffelman, O.: Building a parallel
  computer system for \$18,000 that performs a half peta-flop per day.
\newblock In: Proceedings of the 1st Annual Conference on Genetic and
  Evolutionary Computation-Volume 2, vol.~2, pp. 1484--1490 (1999)

\bibitem{blumofe1996cilk}
Blumofe, R.D., Joerg, C.F., Kuszmaul, B.C., Leiserson, C.E., Randall, K.H.,
  Zhou, Y.: {Cilk}: An efficient multithreaded runtime system.
\newblock Journal of Parallel and Distributed Computing \textbf{37}(1), 55--69
  (1996).
\newblock \doi{10.1006/jpdc.1996.0107}

\bibitem{bocquet2018memory}
Bocquet, M., Hirztlin, T., Klein, J.O., Nowak, E., Vianello, E., Portal, J.M.,
  Querlioz, D.: In-memory and error-immune differential {RRAM} implementation
  of binarized deep neural networks.
\newblock In: 2018 IEEE International Electron Devices Meeting (IEDM) (2018).
\newblock \doi{10.1109/iedm.2018.8614639}

\bibitem{Boerner2023}
Boerner, T.J., Deems, S., Furlani, T.R., Knuth, S.L., Towns, J.: {ACCESS}:
  Advancing innovation: {NSF}'s advanced cyberinfrastructure coordination
  ecosystem: Services \& support.
\newblock In: Practice and Experience in Advanced Research Computing, pp.
  173--176 (2023).
\newblock \doi{10.1145/3569951.3597559}

\bibitem{bonabeau1994we}
Bonabeau, E.W., Theraulaz, G.: Why {ALife}?
\newblock Art. Life \textbf{1}(3), 303--325 (1994).
\newblock \doi{10.1162/artl.1994.1.3.303}

\bibitem{Brashear2025}
Brashear, W., Chakravorty, D., He, Z., O'Connor, D., Siegmann, E., Buitrago,
  P.A., Sanielevici, S.: Byteboost: An advanced cybertraining program designed
  to enhance research on testbed systems.
\newblock In: Practice and Experience in Advanced Research Computing 2025: The
  Power of Collaboration, pp. 1--5 (2025).
\newblock \doi{10.1145/3708035.3736082}

\bibitem{buitrago2021neocortex}
Buitrago, P.A., Nystrom, N.A.: Neocortex and Bridges-2: A High Performance
  {AI}+{HPC} Ecosystem for Science, Discovery, and Societal Good, pp. 205--219
  (2021).
\newblock \doi{10.1007/978-3-030-68035-0_15}

\bibitem{Buluc2021}
Buluc, A., Kolda, T., Wild, S., Anitescu, M., Degennaro, A., Jakeman, J.,
  Kamath, C., Kannan, R., Lopes, M., Martinsson, P.G., Myers, K., Nelson, J.,
  Restrepo, J., Seshadri, C., Vrabie, D., Wohlberg, B., Wright, S., Yang, C.,
  Zwart, P.: Randomized Algorithms for Scientific Computing (RASC) (2021).
\newblock \doi{10.2172/1807223}

\bibitem{bundy2021footprint}
Bundy, J., Ofria, C., Lenski, R.E.: How the footprint of history shapes the
  evolution of digital organisms.
\newblock bioRxiv  (2021).
\newblock \doi{10.1101/2021.04.29.442046}

\bibitem{burke2003increased}
Burke, E., Gustafson, S., Kendall, G., Krasnogor, N.: Is increased diversity in
  genetic programming beneficial? an analysis of lineage selection.
\newblock In: The 2003 Congress on Evolutionary Computation, 2003. CEC '03.,
  vol.~2. {IEEE} (2003).
\newblock \doi{10.1109/cec.2003.1299834}

\bibitem{burlacu2015methods}
Burlacu, B., Affenzeller, M., Winkler, S., Kommenda, M., Kronberger, G.:
  Methods for Genealogy and Building Block Analysis in Genetic Programming, pp.
  61--74.
\newblock Springer International Publishing (2015).
\newblock \doi{10.1007/978-3-319-15720-7_5}

\bibitem{casti1997wouldbe}
Casti, J.L.: Would-Be Worlds: How Simulation is Changing the Frontiers of
  Science.
\newblock John Wiley \& Sons, New York (1997)

\bibitem{chakradhar2010best}
Chakradhar, S.T., Raghunathan, A.: Best-effort computing: Re-thinking parallel
  software and hardware.
\newblock In: Design Automation Conference. {IEEE} (2010).
\newblock \doi{10.1145/1837274.1837492}

\bibitem{chakrapani2008probabilistic}
Chakrapani, L.N., Korkmaz, P., Akgul, B.E., Palem, K.V.: Probabilistic
  system-on-a-chip architectures.
\newblock ACM Transactions on Design Automation of Electronic Systems (TODAES)
  \textbf{12}(3) (2008).
\newblock \doi{10.1145/1255456.1255466}

\bibitem{chamberlain2007parallel}
Chamberlain, B.L., Callahan, D., Zima, H.P.: Parallel programmability and the
  {Chapel} language.
\newblock The International Journal of High Performance Computing Applications
  \textbf{21}(3) (2007).
\newblock \doi{10.1177/1094342007078442}

\bibitem{channon2019maximum}
Channon, A.: Maximum individual complexity is indefinitely scalable in {Geb}.
\newblock Artificial Life \textbf{25}(2), 134--144 (2019).
\newblock \doi{10.1162/artl_a_00285}

\bibitem{chippa2014scalable}
Chippa, V.K., Mohapatra, D., Roy, K., Chakradhar, S.T., Raghunathan, A.:
  Scalable effort.
\newblock IEEE Transactions on Very Large Scale Integration (VLSI) Systems
  \textbf{22}(9) (2014).
\newblock \doi{10.1109/tvlsi.2013.2276759}

\bibitem{cho2012ersa}
Cho, H., Leem, L., Mitra, S.: {ERSA}: Error resilient system architecture for
  probabilistic applications.
\newblock IEEE Transactions on Computer-Aided Design of Integrated Circuits and
  Systems \textbf{31}(4), 546--558 (2012).
\newblock \doi{10.1109/tcad.2011.2179038}

\bibitem{covert2013experiments}
Covert, A.W., Lenski, R.E., Wilke, C.O., Ofria, C.: Experiments on the role of
  deleterious mutations as stepping stones in adaptive evolution.
\newblock Proceedings of the National Academy of Sciences \textbf{110}(34),
  e3171--e3178 (2013).
\newblock \doi{10.1073/pnas.1313424110}

\bibitem{daudey2024aevol}
Daudey, H., Parsons, D.P., Tannier, E., Daubin, V., Boussau, B., Liard, V.,
  Gall\'{e}, R., Rouzaud-Cornabas, J., Beslon, G.: Aevol\_4b: Bridging the gap
  between artificial life and bioinformatics.
\newblock In: The 2024 Conference on Artificial Life. MIT Press (2024).
\newblock \doi{10.1162/isal_a_00716}

\bibitem{dawkins2016ancestor}
Dawkins, R., Wong, Y.: The Ancestor's Tale: A Pilgrimage to the Dawn of
  Evolution.
\newblock Mariner Books, Boston, MA (2016)

\bibitem{togopic2019_rackserver}
{DBCLS TogoTV}: Rack-optimised servers (2019).
\newblock \doi{10.7875/togopic.2019.35}.
\newblock Designed by Watanabe

\bibitem{togopic2024_supercomputer}
{DBCLS TogoTV}: Supercomputer system (2024).
\newblock \doi{10.7875/togopic.2024.193}.
\newblock Designed by erico

\bibitem{dean2012large}
Dean, J., Corrado, G., Monga, R., Chen, K., Devin, M., Mao, M., Ranzato, M.a.,
  Senior, A., Tucker, P., Yang, K., Le, Q., Ng, A.: Large scale distributed
  deep networks.
\newblock In: Advances in Neural Information Processing Systems, vol.~25 (2012)

\bibitem{desell2009robust}
Desell, T., Magdon-Ismail, M., Szymanski, B., Varela, C., Newberg, H., Cole,
  N.: Robust asynchronous optimization for volunteer computing grids.
\newblock In: 2009 Fifth IEEE International Conference on e-Science, pp.
  263--270 (2009).
\newblock \doi{10.1109/e-science.2009.44}

\bibitem{dolson2020interpreting}
Dolson, E., Lalejini, A., Jorgensen, S., Ofria, C.: Interpreting the tape of
  life: Ancestry-based analyses provide insights and intuition about
  evolutionary dynamics.
\newblock Artificial Life \textbf{26}(1), 58--79 (2020).
\newblock \doi{10.1162/artl_a_00313}

\bibitem{dolson2017spatial}
Dolson, E., Ofria, C.: Spatial resource heterogeneity creates local hotspots of
  evolutionary potential.
\newblock In: ECAL 2017, the Fourteenth European Conference on Artificial Life
  (2017).
\newblock \doi{10.1162/isal_a_023}

\bibitem{dolson2021digital}
Dolson, E., Ofria, C.: Digital evolution for ecology research: A review.
\newblock Frontiers in Ecology and Evolution \textbf{9} (2021).
\newblock \doi{10.3389/fevo.2021.750779}

\bibitem{dolson2023phylotrackpy}
Dolson, E., Rodriguez-Papa, S., Moreno, M.A.: Phylotrack: {C++} and {Python}
  libraries for in silico phylogenetic tracking  (2024).
\newblock \doi{10.48550/arXiv.2405.09389}

\bibitem{dongarra2014applied}
Dongarra, J., Hittinger, J., Bell, J., Chacon, L., Falgout, R., Heroux, M.,
  Hovland, P., Ng, E., Webster, C., Wild, S.: Applied mathematics research for
  exascale computing.
\newblock Tech. rep., Lawrence Livermore National Lab.(LLNL) (2014).
\newblock \doi{10.2172/1149042}

\bibitem{dreslinski2010near}
Dreslinski, R.G., Wieckowski, M., Blaauw, D., Sylvester, D., Mudge, T.:
  Near-threshold computing: Reclaiming moore's law through energy efficient
  integrated circuits.
\newblock Proceedings of the IEEE \textbf{98}(2) (2010).
\newblock \doi{10.1109/jproc.2009.2034764}

\bibitem{el2006upc}
El-Ghazawi, T., Smith, L.: {UPC}: Unified parallel {C}.
\newblock In: Proceedings of the 2006 ACM/IEEE Conference on Supercomputing,
  pp. 27--es. Association for Computing Machinery (2006).
\newblock \doi{10.1145/1188455.1188483}

\bibitem{emani2021accelerating}
Emani, M., Vishwanath, V., Adams, C., Papka, M.E., Stevens, R., Florescu, L.,
  Jairath, S., Liu, W., Nama, T., Sujeeth, A.: Accelerating scientific
  applications with sambanova reconfigurable dataflow architecture.
\newblock Computing in Science \& Engineering \textbf{23}(2), 114--119 (2021).
\newblock \doi{10.1109/mcse.2021.3057203}

\bibitem{faithConservationEvaluationPhylogenetic1992}
Faith, D.P.: Conservation evaluation and phylogenetic diversity.
\newblock Biological Conservation \textbf{61}(1), 1--10 (1992).
\newblock \doi{10.1016/0006-3207(92)91201-3}

\bibitem{felsenstein1985phylogenies}
Felsenstein, J.: Phylogenies and the comparative method.
\newblock The American Naturalist \textbf{125}(1), 1--15 (1985).
\newblock \doi{10.1086/284325}

\bibitem{ferguson2023potentiating}
Ferguson, A.J., et~al.: Potentiating mutations facilitate the evolution of
  associative learning in digital organisms.
\newblock In: The 2023 Conference on Artificial Life (2023).
\newblock \doi{10.1162/isal_a_00684}

\bibitem{fortuna2019coevolutionary}
Fortuna, M.A., et~al.: Coevolutionary dynamics shape the structure of
  bacteria-phage infection networks.
\newblock Evolution \textbf{73}(5), 1001--1011 (2019).
\newblock \doi{10.1111/evo.13731}

\bibitem{frenchHostPhylogenyShapes2023}
French, R.K., Anderson, S.H., Cain, K.E., Greene, T.C., Minor, M., Miskelly,
  C.M., Montoya, J.M., Wille, M., Muller, C.G., Taylor, M.W., Digby, A., Crane,
  J., Davitt, G., Eason, D., Hedman, P., Jeynes, B., Latimer, S., Little, S.,
  Mitchell, M., Osborne, J., Philp, B., Salton, A., Uddstrom, L., Vercoe, D.,
  Webster, A., Holmes, E.C.: Host phylogeny shapes viral transmission networks
  in an island ecosystem.
\newblock Nature Ecology \& Evolution \textbf{7}(11), 1834--1843 (2023).
\newblock \doi{10.1038/s41559-023-02192-9}

\bibitem{frontiere2025cosmological}
Frontiere, N., et~al.: Cosmological hydrodynamics at exascale.
\newblock In: Supercomputing, pp. 25--35. ACM (2025).
\newblock \doi{10.1145/3712285.3771786}

\bibitem{gabor2017genealogical}
Gabor, T., Belzner, L.: Genealogical distance in {EA}.
\newblock {GECCO} '17, pp. 1572--1577. ACM (2017).
\newblock \doi{10.1145/3067695.3082529}

\bibitem{gabor2018inheritancebased}
Gabor, T., Belzner, L., Linnhoff-Popien, C.: Inheritance-based convergence
  control in {EA}.
\newblock {GECCO} '18, pp. 841--848. ACM (2018).
\newblock \doi{10.1145/3205455.3205630}

\bibitem{gabor2018preparing}
Gabor, T., Belzner, L., Phan, T., Schmid, K.: Preparing for the unexpected:
  Diversity improves planning resilience in evolutionary algorithms.
\newblock In: 2018 IEEE International Conference on Autonomic Computing
  ({ICAC}), pp. 131--140. IEEE (2018).
\newblock \doi{10.1109/icac.2018.00023}

\bibitem{ganapati2009modular}
Ganapati, P.: Hardware hackers create a modular motherboard.
\newblock \url{wired.com/2009/08/modular-motherboard} (2009).
\newblock Gadget Lab blog, Wired.com; accessed 2026-05-27

\bibitem{genthon2023cell}
Genthon, A., Nozoe, T., Peliti, L., Lacoste, D.: Cell lineage statistics with
  incomplete population trees.
\newblock PRX Life \textbf{1}(1) (2023).
\newblock \doi{10.1103/prxlife.1.013014}

\bibitem{gholami2024ai}
Gholami, A., Yao, Z., Kim, S., Hooper, C., Mahoney, M.W., Keutzer, K.: Memory
  wall.
\newblock IEEE Micro pp. 1--5 (2024).
\newblock \doi{10.1109/mm.2024.3373763}

\bibitem{godindubois2024apoget}
Godin-Dubois, K., Cussat-Blanc, S., Duthen, Y.: {APOGeT} (2024).
\newblock \doi{10.48550/ARXIV.2407.21412}

\bibitem{goldsby2014evolutionary}
Goldsby, H.J., Knoester, D.B., Ofria, C., Kerr, B.: The evolutionary origin of
  somatic cells under the dirty work hypothesis.
\newblock PLoS Biology \textbf{12}(5), e1001858 (2014).
\newblock \doi{10.1371/journal.pbio.1001858}

\bibitem{grabowski2013case}
Grabowski, L.M., Bryson, D.M., Dyer, F.C., Pennock, R.T., Ofria, C.: A case
  study of the de novo evolution of a complex odometric behavior in digital
  organisms.
\newblock PLoS One \textbf{8}(4), e60466 (2013).
\newblock \doi{10.1371/journal.pone.0060466}

\bibitem{gropp1996high}
Gropp, W., Lusk, E., Doss, N., Skjellum, A.: A high-performance, portable
  implementation of the {MPI} message passing interface standard.
\newblock Parallel Computing \textbf{22}(6), 789--828 (1996).
\newblock \doi{10.1016/0167-8191(96)00024-5}

\bibitem{grundler2025geographic}
Grundler, M.C., Terhorst, J., Bradburd, G.S.: A geographic history of human
  genetic ancestry.
\newblock Science \textbf{387}(6741) (2025).
\newblock \doi{10.1126/science.adp4642}

\bibitem{guijt2023impact}
Guijt, A., Thierens, D., Alderliesten, T., Bosman, P.A.: The impact of
  asynchrony on parallel model-based eas.
\newblock In: Proceedings of the Genetic and Evolutionary Computation
  Conference, pp. 910--918 (2023).
\newblock \doi{10.1145/3583131.3590406}

\bibitem{gunther2014algorithm}
Gunther, J.C.: Compressing circular buffers.
\newblock Acm Tms \textbf{40}(2), 1--12 (2014).
\newblock \doi{10.1145/2559995}

\bibitem{gupta2017failures}
Gupta, S., Patel, T., Engelmann, C., Tiwari, D.: Failures in large scale
  systems: long-term measurement, analysis, and implications.
\newblock In: Proceedings of the International Conference for High Performance
  Computing, Networking, Storage and Analysis, pp. 1--12 (2017).
\newblock \doi{10.1145/3126908.3126937}

\bibitem{haller2023slim}
Haller, B.C., Messer, P.W.: Slim 4: Multispecies eco-evolutionary modeling.
\newblock The American Naturalist \textbf{201}(5), E127--e139 (2023).
\newblock \doi{10.1086/723601}

\bibitem{hamming1950error}
Hamming, R.W.: Error detecting and error correcting codes.
\newblock Bell System Technical Journal \textbf{29}(2), 147--160 (1950).
\newblock \doi{10.1002/j.1538-7305.1950.tb00463.x}

\bibitem{harada2022frequency}
Harada, T.: A frequency-based parent selection for reducing the effect of
  evaluation time bias in asynchronous parallel multi-objective evolutionary
  algorithms.
\newblock Natural Computing \textbf{24}(2), 211--225 (2022).
\newblock \doi{10.1007/s11047-022-09940-z}

\bibitem{harris2020array}
Harris, C.R., Millman, K.J., van~der Walt, S.J., Gommers, R., Virtanen, P.,
  Cournapeau, D., Wieser, E., Taylor, J., Berg, S., Smith, N.J., Kern, R.,
  Picus, M., Hoyer, S., van Kerkwijk, M.H., Brett, M., Haldane, A., del
  R{\'{i}}o, J.F., Wiebe, M., Peterson, P., G{\'{e}}rard-Marchant, P.,
  Sheppard, K., Reddy, T., Weckesser, W., Abbasi, H., Gohlke, C., Oliphant,
  T.E.: Array programming with {NumPy}.
\newblock Nature \textbf{585}(7825), 357--362 (2020).
\newblock \doi{10.1038/s41586-020-2649-2}

\bibitem{hedges2015tree}
Hedges, S.B., Marin, J., Suleski, M., Paymer, M., Kumar, S.: Tree of life
  reveals clock-like speciation and diversification.
\newblock Molecular Biology and Evolution \textbf{32}(4), 835--845 (2015).
\newblock \doi{10.1093/molbev/msv037}

\bibitem{hernandez2022can}
Hernandez, J.G., Lalejini, A., Dolson, E.: What {{Can Phylogenetic Metrics
  Tell}} us {{About Useful Diversity}} in {{Evolutionary Algorithms}}?
\newblock In: Genetic {{Programming Theory}} and {{Practice XVIII}}, pp.
  63--82. Springer Nature (2022).
\newblock \doi{10.1007/978-981-16-8113-4_4}

\bibitem{hernandez2022dossier}
Hernandez, J.G., Lalejini, A., Ofria, C.: A suite of diagnostic metrics for
  characterizing selection schemes (2022).
\newblock \doi{10.48550/arxiv.2204.13839}

\bibitem{heroux2014toward}
Heroux, M.A.: Toward resilient algorithms and applications (2014).
\newblock \doi{10.48550/arXiv.1402.3809}

\bibitem{hinchliff2015synthesis}
Hinchliff, C.E., Smith, S.A., Allman, J.F., Burleigh, J.G., Chaudhary, R.,
  Coghill, L.M., Crandall, K.A., Deng, J., Drew, B.T., Gazis, R., Gude, K.,
  Hibbett, D.S., Katz, L.A., Laughinghouse, H.D., McTavish, E.J., Midford,
  P.E., Owen, C.L., Ree, R.H., Rees, J.A., Soltis, D.E., Williams, T.,
  Cranston, K.A.: Synthesis of phylogeny and taxonomy into a comprehensive tree
  of life.
\newblock Proceedings of the National Academy of Sciences \textbf{112}(41)
  (2015).
\newblock \doi{10.1073/pnas.1423041112}

\bibitem{Hooker1995}
Hooker, J.N.: Testing heuristics: We have it all wrong.
\newblock J. of Heur. \textbf{1}(1), 33--42 (1995).
\newblock \doi{10.1007/bf02430364}

\bibitem{huberman1993evolutionary}
Huberman, B.A., Glance, N.S.: Evolutionary games and computer simulations.
\newblock Proceedings of the National Academy of Sciences \textbf{90}(16),
  7716--7718 (1993).
\newblock \doi{10.1073/pnas.90.16.7716}

\bibitem{hunter2007matplotlib}
Hunter, J.D.: {Matplotlib}: A {2D} graphics environment.
\newblock Computing in Science \& Engineering \textbf{9}(3), 90--95 (2007).
\newblock \doi{10.1109/mcse.2007.55}

\bibitem{izzo2009parallel}
Izzo, D., Rucinski, M., Ampatzis, C.: Parallel global optimisation
  meta-heuristics using an asynchronous island-model.
\newblock In: 2009 IEEE Congress on Evolutionary Computation, pp. 2301--2308
  (2009).
\newblock \doi{10.1109/cec.2009.4983227}

\bibitem{jia2019dissecting}
Jia, Z., Tillman, B., Maggioni, M., Scarpazza, D.P.: Dissecting the {Graphcore}
  {IPU} architecture via microbenchmarking.
\newblock arXiv  (2019).
\newblock \doi{10.48550/arxiv.1912.03413}

\bibitem{jones2020inference}
Jones, M.G., Khodaverdian, A., Quinn, J.J., Chan, M.M., Hussmann, J.A., Wang,
  R., Xu, C., Weissman, J.S., Yosef, N.: Inference of single-cell phylogenies
  from lineage tracing data using {Cassiopeia}.
\newblock Genome Biology \textbf{21}(1) (2020).
\newblock \doi{10.1186/s13059-020-02000-8}

\bibitem{jones2024spatiotemporal}
Jones, M.G., et~al.: Spatiotemporal lineage tracing reveals the dynamic spatial
  architecture of tumour growth and metastasis.
\newblock {openRxiv}  (2024).
\newblock \doi{10.1101/2024.10.21.619529}

\bibitem{kale1993charm++}
Kale, L.V., Krishnan, S.: {Charm++} a portable concurrent object oriented
  system based on {C++}.
\newblock In: Proceedings of the Eighth Annual Conference on Object-oriented
  Programming Systems, Languages, and Applications, pp. 91--108 (1993).
\newblock \doi{10.1145/165854.165874}

\bibitem{kaplan2020scaling}
Kaplan, J., McCandlish, S., Henighan, T., Brown, T.B., Chess, B., Child, R.,
  Gray, S., Radford, A., Wu, J., Amodei, D.: Scaling laws for neural language
  models (2020).
\newblock \doi{10.48550/arxiv.2001.08361}

\bibitem{karakus2017quality}
Karakus, M., Durresi, A.: Quality of service ({QoS}) in software defined
  networking ({SDN}): A survey.
\newblock {JNCA} \textbf{80}, 200--218 (2017).
\newblock \doi{10.1016/j.jnca.2016.12.019}

\bibitem{karns2025evaluation}
Karns, J., Desell, T.: Evaluation time bias in asynchronous evolutionary
  algorithms: A replication study and a novel mitigation strategy.
\newblock In: Proceedings of the Genetic and Evolutionary Computation
  Conference, pp. 13--21 (2025).
\newblock \doi{10.1145/3712256.3726458}

\bibitem{kasap2018dynamic}
Kasap, B., van Opstal, A.J.: Dynamic parallelism for synaptic updating in
  {GPU}-accelerated spiking neural network simulations.
\newblock Neurocomputing \textbf{302}, 55--65 (2018).
\newblock \doi{10.1016/j.neucom.2018.04.007}

\bibitem{kashyap2006ip}
Kashyap, V.: {IP} over {InfiniBand} ({IPoIB}) architecture.
\newblock The Internet Society \textbf{22} (2006)

\bibitem{kebschull2018cellular}
Kebschull, J.M., Zador, A.M.: Cellular barcoding: lineage tracing, screening
  and beyond.
\newblock Nature Methods \textbf{15}(11), 871--879 (2018).
\newblock \doi{10.1038/s41592-018-0185-x}

\bibitem{khan2021analysis}
Khan, A., Sim, H., Vazhkudai, S.S., Butt, A.R., Kim, Y.: An analysis of system
  balance and architectural trends based on top500 supercomputers.
\newblock In: The International Conference on High Performance Computing in
  Asia-Pacific Region (2021).
\newblock \doi{10.1145/3432261.3432263}

\bibitem{kim2006discovery}
Kim, T.K., Hewavitharana, A.K., Shaw, P.N., Fuerst, J.A.: Discovery of a new
  source of rifamycin antibiotics in marine sponge actinobacteria by
  phylogenetic prediction.
\newblock Applied and Environmental Microbiology \textbf{72}(3), 2118--2125
  (2006).
\newblock \doi{10.1128/aem.72.3.2118-2125.2006}

\bibitem{kokolis2025revisiting}
Kokolis, A., Kuchnik, M., Hoffman, J., Kumar, A., Malani, P., Ma, F., DeVito,
  Z., Sengupta, S., Saladi, K., Wu, C.J.: Revisiting reliability in large-scale
  machine learning research clusters.
\newblock In: 2025 IEEE International Symposium on High Performance Computer
  Architecture (HPCA), pp. 1259--1274. {IEEE} (2025).
\newblock \doi{10.1109/hpca61900.2025.00096}

\bibitem{konno2022deep}
Konno, N., Kijima, Y., Watano, K., Ishiguro, S., Ono, K., Tanaka, M., Mori, H.,
  Masuyama, N., Pratt, D., Ideker, T., Iwasaki, W., Yachie, N.: Deep
  distributed computing to reconstruct extremely large lineage trees.
\newblock Nature Biotechnology \textbf{40}(4), 566--575 (2022).
\newblock \doi{10.1038/s41587-021-01111-2}

\bibitem{koop2007high}
Koop, M.J., Sur, S., Gao, Q., Panda, D.K.: High performance {MPI} design using
  unreliable datagram for ultra-scale {InfiniBand} clusters.
\newblock In: Proceedings of the 21st Annual International Conference on
  Supercomputing (2007).
\newblock \doi{10.1145/1274971.1274997}

\bibitem{la2020cerebras}
La, M., Chien, A.: {Cerebras} systems: Journey to the wafer-scale engine.
\newblock University of Chicago, Tech. Rep  (2020)

\bibitem{lalejini2024phylogeny}
Lalejini, A., Moreno, M.A., Hernandez, J.G., Dolson, E.: Phylogeny-informed
  fitness estimation for test-based parent selection.
\newblock In: Genetic Programming Theory and Practice XX, pp. 241--261.
  Springer International Publishing (2024).
\newblock \doi{10.1007/978-981-99-8413-8_13}

\bibitem{lalejini2018evolving}
Lalejini, A., Ofria, C.: Evolving event-driven programs with {SignalGP}.
\newblock In: Proceedings of the Genetic and Evolutionary Computation
  Conference (2018).
\newblock \doi{10.1145/3205455.3205523}

\bibitem{lalejini2024runtime}
Lalejini, A., Sanson, M., Garbus, J., Moreno, M.A., Dolson, E.: Runtime
  phylogenetic analysis enables extreme subsampling for test-based problems.
\newblock In: Proceedings of the Genetic and Evolutionary Computation
  Conference Companion, pp. 511--514. Association for Computing Machinery
  (2024).
\newblock \doi{10.1145/3638530.3654208}

\bibitem{Landenmark2015}
Landenmark, H.K.E., Forgan, D.H., Cockell, C.S.: An estimate of the total {DNA}
  in the biosphere.
\newblock {PLOS} Biology \textbf{13}(6), e1002168 (2015).
\newblock \doi{10.1371/journal.pbio.1002168}

\bibitem{Lehman2020}
Lehman, J., Clune, J., Misevic, D., Adami, C., Altenberg, L., Beaulieu, J.,
  Bentley, P.J., Bernard, S., Beslon, G., Bryson, D.M., Cheney, N., Chrabaszcz,
  P., Cully, A., Doncieux, S., Dyer, F.C., Ellefsen, K.O., Feldt, R., Fischer,
  S., Forrest, S., F\'{r}enoy, A., Gag\'{n}e, C., Le~Goff, L., Grabowski, L.M.,
  Hodjat, B., Hutter, F., Keller, L., Knibbe, C., Krcah, P., Lenski, R.E.,
  Lipson, H., MacCurdy, R., Maestre, C., Miikkulainen, R., Mitri, S., Moriarty,
  D.E., Mouret, J.B., Nguyen, A., Ofria, C., Parizeau, M., Parsons, D.,
  Pennock, R.T., Punch, W.F., Ray, T.S., Schoenauer, M., Schulte, E., Sims, K.,
  Stanley, K.O., Taddei, F., Tarapore, D., Thibault, S., Watson, R., Weimer,
  W., Yosinski, J.: The surprising creativity of digital evolution: A
  collection of anecdotes from the evolutionary computation and artificial life
  research communities.
\newblock Art. L. \textbf{26}(2), 274--306 (2020).
\newblock \doi{10.1162/artl_a_00319}

\bibitem{leith2012wlan}
Leith, D.J., Clifford, P., Badarla, V., Malone, D.: {WLAN} channel selection
  without communication.
\newblock Computer Networks \textbf{56}(4) (2012).
\newblock \doi{10.1016/j.comnet.2011.12.015}

\bibitem{lemey2010phylogeography}
Lemey, P., Rambaut, A., Welch, J.J., Suchard, M.A.: Phylogeography takes a
  relaxed random walk in continuous space and time.
\newblock Molecular Biology and Evolution \textbf{27}(8), 1877--1885 (2010).
\newblock \doi{10.1093/molbev/msq067}

\bibitem{lenski2003evolutionary}
Lenski, R.E., Ofria, C., Pennock, R.T., Adami, C.: The evolutionary origin of
  complex features.
\newblock Nature \textbf{423}(6936), 139--144 (2003).
\newblock \doi{10.1038/nature01568}

\bibitem{levy2015quantitative}
Levy, S.F., Blundell, J.R., Venkataram, S., Petrov, D.A., Fisher, D.S.,
  Sherlock, G.: Quantitative evolutionary dynamics using high-resolution
  lineage tracking.
\newblock Nature \textbf{519}(7542) (2015).
\newblock \doi{10.1038/nature14279}

\bibitem{lewinsohnStatedependentEvolutionaryModels2023a}
Lewinsohn, M.A., Bedford, T., M\"{u}ller, N.F., Feder, A.F.: State-dependent
  evolutionary models reveal modes of solid tumour growth.
\newblock Nature Ecology \& Evolution \textbf{7}(4), 581--596 (2023).
\newblock \doi{10.1038/s41559-023-02000-4}

\bibitem{liang2024asynchronous}
Liang, J., Shahrzad, H., Miikkulainen, R.: Asynchronous evolution of deep
  neural network architectures.
\newblock Applied Soft Computing \textbf{152}, 111,209 (2024).
\newblock \doi{10.1016/j.asoc.2023.111209}

\bibitem{lyu2021improving}
Lyu, Z., Karns, J., ElSaid, A., Mkaouer, M., Desell, T.: Improving Distributed
  Neuroevolution Using Island Extinction and Repopulation, pp. 568--583 (2021).
\newblock \doi{10.1007/978-3-030-72699-7_36}

\bibitem{lyu2023online}
Lyu, Z., Ororbia, A., Desell, T.: Online evolutionary neural architecture
  search for multivariate non-stationary time series forecasting.
\newblock Applied Soft Computing \textbf{145}, 110,522 (2023).
\newblock \doi{10.1016/j.asoc.2023.110522}

\bibitem{marcus2018deep}
Marcus, G.: Deep learning: A critical appraisal (2018).
\newblock \doi{10.48550/arxiv.1801.00631}

\bibitem{markovi2020physics}
Markovi\'{c}, D., Mizrahi, A., Querlioz, D., Grollier, J.: Neur. comp.
\newblock Nature Reviews Physics \textbf{2}(9), 499--510 (2020).
\newblock \doi{10.1038/s42254-020-0208-2}

\bibitem{Masuyama2019}
Masuyama, N., Mori, H., Yachie, N.: {DNA} barcodes evolve for high-resolution
  cell lineage tracing.
\newblock Current Opinion in Chemical Biology \textbf{52} (2019).
\newblock \doi{10.1016/j.cbpa.2019.05.014}

\bibitem{mayr2019spinnaker210million}
Mayr, C., Hoeppner, S., Furber, S.: {SpiNNaker} 2: A 10 million core processor
  system for brain simulation and machine learning (2019).
\newblock \doi{10.48550/arXiv.1911.02385}

\bibitem{mckinney-proc-scipy-2010}
{M}c{K}inney: {D}ata {S}tructures for {S}tatistical {C}omputing in {P}ython.
\newblock In: {P}roceedings of the 9th {P}ython in {S}cience {C}onference, pp.
  56--61 (2010).
\newblock \doi{10.25080/Majora-92bf1922-00a}

\bibitem{mcphee2016using}
McPhee, N.F., Donatucci, D., Helmuth, T.: Using Graph Databases to Explore the
  Dynamics of Genetic Programming Runs, pp. 185--201.
\newblock Springer International Publishing (2016).
\newblock \doi{10.1007/978-3-319-34223-8_11}

\bibitem{medina2020habana}
Medina, E., Dagan, E.: Habana labs purpose-built ai inference and training
  processor architectures: Scaling ai training systems using standard ethernet
  with gaudi processor.
\newblock IEEE Micro \textbf{40}(2), 17--24 (2020).
\newblock \doi{10.1109/mm.2020.2975185}

\bibitem{meng2009best}
Meng, J., Chakradhar, S., Raghunathan, A.: Best-effort parallel execution
  framework for recognition and mining applications.
\newblock In: 2009 IEEE International Symposium on Parallel \& Distributed
  Processing, pp. 1--12. {IEEE} (2009).
\newblock \doi{10.1109/ipdps.2009.5160991}

\bibitem{menon2025reproducibility}
Menon, H.: Reproducibility in the age of approximate computing.
\newblock Better Scientific Software Blog  (2025)

\bibitem{menon2023approximate}
Menon, H., Diffenderfer, J., Georgakoudis, G., Laguna, I., Lam, M.O.,
  Osei-Kuffuor, D., Parasyris, K., Vanover, J.: Approximate high-performance
  computing: A fast and energy-efficient computing paradigm in the post-moore
  era.
\newblock IT Prof. \textbf{25}(2), 7--15 (2023).
\newblock \doi{10.1109/mitp.2023.3254642}

\bibitem{mittal2016survey}
Mittal: A survey of techniques for approximate computing.
\newblock ACM Computing Surveys (CSUR) \textbf{48}(4), 1--33 (2016).
\newblock \doi{10.1145/2893356}

\bibitem{moreno_2026_trust}
Moreno: Trust but verify supplement (2026).
\newblock \doi{10.17605/osf.io/pcu9x}.
\newblock \urlprefix\url{osf.io/pcu9x}

\bibitem{moreno2023toward}
Moreno, Dolson, Rodriguez-Papa: Toward phylogenetic inference of evolutionary
  dynamics at scale.
\newblock In: Alife 2023, p.~79. {MIT} Press (2023).
\newblock \doi{10.1162/isal_a_00694}

\bibitem{moreno2022engineering}
Moreno, M.A.: Engineering scalable digital models to study major transitions in
  evolution.
\newblock Ph.D. thesis, Michigan State University, East Lansing, MI (2022)

\bibitem{moreno2024methods}
Moreno, M.A.: Methods for rich phylogenetic inference over distributed sexual
  populations.
\newblock In: Genetic Programming Theory and Practice XX, pp. 125--141.
  Springer International Publishing (2024).
\newblock \doi{10.1007/978-981-99-8413-8_7}

\bibitem{moreno2022hereditary}
Moreno, M.A., Dolson, E., Ofria, C.: Hereditary stratigraphy: Genome
  annotations to enable phylogenetic inference over distributed populations.
\newblock In: {ALife}, p.~64. {MIT} Press (2022).
\newblock \doi{10.1162/isal_a_00550}

\bibitem{moreno2022hstrat}
Moreno, M.A., Dolson, E., Ofria, C.: hstrat: a {Python} package for
  phylogenetic inference on distributed digital evolution populations.
\newblock Journal of Open Source Software \textbf{7}(80), 4866 (2022).
\newblock \doi{10.21105/joss.04866}

\bibitem{moreno2025hypermutatorsc}
Moreno, M.A., Dolson, E., Zaman, L.: Wafer-scale simulation of mutator allele
  dynamics in large asexual populations.
\newblock In: SC25 Research Poster and ACM Student Research Competition Poster
  Archive (2025)

\bibitem{moreno2025extending}
Moreno, M.A., Fard, S.H., Zaman, L., Dolson, E.: Extending a phylogeny-based
  method for detecting signatures of multi-level selection for applications in
  artificial life.
\newblock In: ALIFE 2025: Ciphers of Life: Proceedings of the Artificial Life
  Conference 2025, vol.~37. MIT Press (2025).
\newblock \doi{10.1162/isal.a.916}

\bibitem{moreno2019toward}
Moreno, M.A., Ofria, C.: Toward open-ended fraternal transitions in
  individuality.
\newblock Artificial Life \textbf{25}(2), 117--133 (2019).
\newblock \doi{10.1162/artl_a_00284}

\bibitem{moreno2021exploring}
Moreno, M.A., Ofria, C.: Exploring evolved multicellular life histories in a
  open-ended digital evolution system.
\newblock Frontiers in Ecology and Evolution \textbf{10} (2022).
\newblock \doi{10.3389/fevo.2022.750837}

\bibitem{moreno2024algorithms}
Moreno, M.A., Papa, S.R., Dolson, E.: Algorithms for efficient, compact online
  data stream curation.
\newblock arXiv  (2024).
\newblock \doi{10.48550/arxiv.2403.00266}

\bibitem{moreno2024analysis}
Moreno, M.A., Papa, S.R., Dolson, E.: Analysis of phylogeny tracking algorithms
  for serial and multiprocess applications.
\newblock arXiv  (2024).
\newblock \doi{10.48550/arxiv.2403.00246}

\bibitem{moreno2021case}
Moreno, M.A., Papa, S.R., Ofria, C.: Case study of novelty, complexity, and
  adaptation in a multicellular system.
\newblock In: {OEE4}: The Fourth Workshop on Open-Ended Evolution (2021).
\newblock
  \urlprefix\url{http://workshops.alife.org/oee4/papers/moreno-oee4-camera-ready.pdf}

\bibitem{moreno2025testing}
Moreno, M.A., Ranjan, A., Dolson, E., Zaman, L.: Testing the inference accuracy
  of accelerator-friendly approximate phylogeny tracking.
\newblock In: 2025 IEEE Symposium on Computational Intelligence in Artificial
  Life and Cooperative Intelligent Systems (ALIFE-CIS), pp. 1--9. {IEEE}
  (2025).
\newblock \doi{10.1109/alife-cis64968.2025.10979833}

\bibitem{Moreno2025ecology}
Moreno, M.A., Rodriguez-Papa, S., Dolson, E.: Ecology, spatial structure, and
  selection pressure induce strong signatures in phylogenetic structure.
\newblock Artificial Life \textbf{31}(2), 129--152 (2025).
\newblock \doi{10.1162/artl_a_00470}

\bibitem{moreno2021signalgp}
Moreno, M.A., {Rodriguez Papa}, S., Lalejini, A., Ofria, C.: {SignalGP}-lite:
  Event driven genetic programming library for large-scale artificial life
  applications (2021).
\newblock \doi{10.48550/arxiv.2108.00382}

\bibitem{moreno2021conduit}
Moreno, M.A., Rodriguez~Papa, S., Ofria, C.: Conduit: A {C++} library for
  best-effort high performance computing.
\newblock In: Proceedings of the Genetic and Evolutionary Computation
  Conference Companion (2021).
\newblock \doi{10.1145/3449726.3463205}

\bibitem{moreno2026phyloframe}
Moreno, M.A., Sukumaran, J., Zaman, L., Dolson, E.: Phyloframe: A
  dataframe-based library for fast, flexible phylogenetic computation (2026).
\newblock \doi{10.48550/arXiv.2605.28545}

\bibitem{moreno2024wse}
Moreno, M.A., Yang, C.: mmore500/wse-async-ga (2024).
\newblock \doi{10.5281/zenodo.16898903}

\bibitem{murphy2008simple}
Murphy, G., Ryan, C.: A simple powerful constraint for genetic programming.
\newblock In: Genetic Programming, pp. 146--157. Springer Berlin Heidelberg
  (2008).
\newblock \doi{10.1007/978-3-540-78671-9_13}

\bibitem{navaux2023challenges}
Navaux, P.O.A., Lorenzon, A.F., Serpa, M.d.S.: Challenges in high-performance
  computing.
\newblock Journal of the Brazilian Computer Society \textbf{29}(1), 51--62
  (2023).
\newblock \doi{10.5753/jbcs.2023.2219}

\bibitem{niu2011hogwild}
Niu, F., Recht, B., Re, C., Wright, S.J.: Hogwild! a lock-free approach to
  parallelizing stochastic gradient descent.
\newblock In: Proceedings of the 24th International Conference on Neural
  Information Processing Systems, pp. 693--701 (2011)

\bibitem{noel2014dogwild}
Noel, C., Osindero, S.: Dogwild!
\newblock In: NeurIPS Workshop on Distributed Machine Learning and Matrix
  Computations, pp. 693--701 (2014)

\bibitem{novembre2008genes}
Novembre, J., Johnson, T., Bryc, K., Kutalik, Z., Boyko, A.R., Auton, A.,
  Indap, A., King, K.S., Bergmann, S., Nelson, M.R., Stephens, M., Bustamante,
  C.D.: Genes mirror geography within europe.
\newblock Nature \textbf{456}(7218), 98--101 (2008).
\newblock \doi{10.1038/nature07331}

\bibitem{nozoe2017inferring}
Nozoe, T., Kussell, E., Wakamoto, Y.: Inferring fitness landscapes and
  selection on phenotypic states from single-cell genealogical data.
\newblock {PLOS} Genetics \textbf{13}(3), e1006653 (2017).
\newblock \doi{10.1371/journal.pgen.1006653}

\bibitem{ofria2004avida}
Ofria, C., Wilke, C.O.: {Avida}: A software platform for research in
  computational evolutionary biology.
\newblock Art. Life \textbf{10}(2), 191--229 (2004).
\newblock \doi{10.1162/106454604773563612}

\bibitem{pan2026integrative}
Pan, X., Chen, Y., Zhang, X.: Integrative inference of spatially resolved cell
  lineage trees using {LineageMap}.
\newblock {openRxiv}  (2026).
\newblock \doi{10.64898/2026.01.19.700383}

\bibitem{pehle2022brainscales}
Pehle, C., Billaudelle, S., Cramer, B., Kaiser, J., Schreiber, K., Stradmann,
  Y., Weis, J., Leibfried, A., M\"{u}ller, E., Schemmel, J.: The
  {BrainScaleS-2} accelerated neuromorphic system with hybrid plasticity.
\newblock Frontiers in Neuroscience \textbf{16} (2022).
\newblock \doi{10.3389/fnins.2022.795876}

\bibitem{penczykowski2016understanding}
Penczykowski, R.M., Laine, A.L., Koskella, B.: Understanding the ecology and
  evolution of host--parasite interactions across scales.
\newblock Evolutionary Applications \textbf{9}(1), 37--52 (2015).
\newblock \doi{10.1111/eva.12294}

\bibitem{pennock2007models}
Pennock, R.T.: Models, simulations, instantiations, and evidence: the case of
  digital evolution.
\newblock Journal of Experimental \& Theoretical Artificial Intelligence
  \textbf{19}(1), 29--42 (2007).
\newblock \doi{10.1080/09528130601116113}

\bibitem{perumalla2022computer}
Perumalla, K., Bremer, M., Brown, K., Chan, C., Eidenbenz, S., Hemmert, K.S.,
  Hoisie, A., Newton, B., Nutaro, J., Oppelstrup, T., et~al.: Computer science
  research needs for parallel discrete event simulation (pdes).
\newblock Tech. rep., Lawrence Livermore National Lab.(LLNL), Livermore, CA
  (United States) (2022).
\newblock \doi{10.2172/1889525}

\bibitem{pontes2020evolutionary}
Pontes, A.C., Mobley, R.B., Ofria, C., Adami, C., Dyer, F.C.: The evolutionary
  origin of associative learning.
\newblock The American Naturalist \textbf{195}(1), e1--e19 (2020).
\newblock \doi{10.1086/706252}

\bibitem{ray1995proposal}
Ray, T.: A proposal to create a network-wide biodiversity reserve for digital
  organisms.
\newblock Tech. Rep. Tr-h-133, Atr, Kyoto, Japan (1995).
\newblock \urlprefix\url{http://tomray.me/pubs/reserves/}

\bibitem{raynes2018sign}
Raynes, Y., Wylie, C.S., Sniegowski, P.D., Weinreich, D.M.: Sign of selection
  on mutation rate modifiers depends on population size.
\newblock Proceedings of the National Academy of Sciences \textbf{115}(13),
  3422--3427 (2018).
\newblock \doi{10.1073/pnas.1715996115}

\bibitem{reinders2007intel}
Reinders, J.: Intel threading building blocks: outfitting {C++} for multi-core
  processor parallelism.
\newblock O'Reilly Media, Inc. (2007)

\bibitem{rhodes2020real}
Rhodes, O., Peres, L., Rowley, A.G., Gait, A., Plana, L.A., Brenninkmeijer, C.,
  Furber, S.B.: Real-time cortical simulation on neuromorphic hardware.
\newblock Philosophical Transactions of the Royal Society A \textbf{378}(2164)
  (2020).
\newblock \doi{10.1098/rsta.2019.0160}

\bibitem{rinard2012unsynchronized}
Rinard, M.: Unsynchronized techniques for approximate parallel computing.
\newblock In: {RACES} Workshop on Relaxing Synchronization for Multicore and
  Manycore Scalability (SPLASH 2012). Tucson, Arizona, USA (2012)

\bibitem{rinard2013parallel}
Rinard, M.: Parallel {Synchronization-Free} approximate data structure
  construction.
\newblock In: 5th USENIX Workshop on Hot Topics in Parallelism (HotPar 13). San
  Jose, CA (2013)

\bibitem{schreiber2021cross}
Schreiber, S.J., Ke, R., Loverdo, C., Park, M., Ahsan, P., Lloyd-Smith, J.O.:
  Cross-scale dynamics and the evolutionary emergence of infectious diseases.
\newblock Virus Evol. \textbf{7}(1) (2021).
\newblock \doi{10.1093/ve/veaa105}

\bibitem{scott2022avoiding}
Scott, E.O., Coletti, M., Schuman, C.D., Kay, B., Kulkarni, S.R., Parsa, M.,
  Gunaratne, C., De~Jong, K.A.: Avoiding excess computation in asynchronous
  evolutionary algorithms.
\newblock {ES} \textbf{40}(5) (2022).
\newblock \doi{10.1111/exsy.13100}

\bibitem{scott2015understanding}
Scott, E.O., De~Jong, K.A.: Understanding simple asynchronous evolutionary
  algorithms.
\newblock In: Proceedings of the 2015 ACM Conference on Foundations of Genetic
  Algorithms XIII, pp. 85--98 (2015).
\newblock \doi{10.1145/2725494.2725509}

\bibitem{scott2016evaluation}
Scott, E.O., De~Jong, K.A.: Evaluation-time bias in quasi-generational and
  steady-state asynchronous evolutionary algorithms.
\newblock In: Proceedings of the Genetic and Evolutionary Computation
  Conference 2016, pp. 845--852 (2016).
\newblock \doi{10.1145/2908812.2908934}

\bibitem{Scott2018}
Scott, J.G., Maini, P.K., Anderson, A.R.A., Fletcher, A.G.: Inferring tumour
  proliferative organisation from phylogenetic tree measures in a computational
  model.
\newblock Systematic Biology \textbf{69}(4), 623--637 (2018).
\newblock \doi{10.1101/334946}

\bibitem{shahbandegan2022untangling}
Shahbandegan, S., Hernandez, J.G., Lalejini, A., Dolson, E.: Untangling
  phylogenetic diversity's role in evolutionary computation using a suite of
  diagnostic fitness landscapes.
\newblock In: Proceedings of the Genetic and Evolutionary Computation
  Conference Companion, pp. 2322--2325. {ACM} (2022).
\newblock \doi{10.1145/3520304.3534028}

\bibitem{singhvi2025scalable}
Singhvi, V., Wagner, J., Dolson, E., Zaman, L., Moreno, M.A.: A scalable trie
  building algorithm for high-throughput phyloanalysis of wafer-scale digital
  evolution experiments.
\newblock In: The 2025 Conference on Artificial Life. MIT Press (2025).
\newblock \doi{10.1162/ISAL.a.890}

\bibitem{stadler2013recovering}
Stadler, T.: Recovering speciation and extinction dynamics based on
  phylogenies.
\newblock Journal of Evolutionary Biology \textbf{26}(6), 1203--1219 (2013).
\newblock \doi{10.1111/jeb.12139}

\bibitem{STAMATAKIS2005phylogenetics}
Stamatakis, A.: Phylogenetics: Applications, software and challenges.
\newblock Cancer Genomics \& Proteomics \textbf{2}(5), 301--305 (2005)

\bibitem{cerebras2025yield}
Systems, C.: 100x defect tolerance: How {Cerebras} solved the yield problem.
\newblock Cerebras Blog (2025)

\bibitem{taylor2019evolutionary}
Taylor, T.: Evolutionary innovations and where to find them: Routes to
  open-ended evolution in natural and artificial systems.
\newblock Art. Life \textbf{25}(2), 207--224 (2019).
\newblock \doi{10.1162/artl_a_00290}

\bibitem{taylor2016open}
Taylor, T., Bedau, M., Channon, A., Ackley, D., Banzhaf, W., Beslon, G.,
  Dolson, E., Froese, T., Hickinbotham, S., Ikegami, T., et~al.: Open-ended
  evolution: Perspectives from the oee workshop in york.
\newblock Art. Life \textbf{22}(3), 408--423 (2016).
\newblock \doi{10.1162/artl_a_00210}

\bibitem{Thompson1997}
Thompson, A.: An evolved circuit, intrinsic in silicon, entwined with physics,
  \emph{LNCS}, vol. 1259 (1997).
\newblock \doi{10.1007/3-540-63173-9_61}

\bibitem{VanEssendelft2025}
Van~Essendelft, D., et~al.: Record acceleration of the two-dimensional ising
  model.
\newblock Computer Physics Communications \textbf{315}, 109,734 (2025).
\newblock \doi{10.1016/j.cpc.2025.109734}

\bibitem{2020SciPy-NMeth}
Virtanen, P., Gommers, R., Oliphant, T.E., Haberland, M., Reddy, T.,
  Cournapeau, D., Burovski, E., Peterson, P., Weckesser, W., Bright, J., {van
  der Walt}, S.J., Brett, M., Wilson, J., Millman, K.J., Mayorov, N., Nelson,
  A.R.J., Jones, E., Kern, R., Larson, E., Carey, C.J., Polat, {\.I}., Feng,
  Y., Moore, E.W., {VanderPlas}, J., Laxalde, D., Perktold, J., Cimrman, R.,
  Henriksen, I., Quintero, E.A., Harris, C.R., Archibald, A.M., Ribeiro, A.H.,
  Pedregosa, F., {van Mulbregt}, P., {SciPy 1.0 Contributors}: {{SciPy} 1.0:
  Fundamental Algorithms for Scientific Computing in {Python}}.
\newblock Nature Methods \textbf{17}, 261--272 (2020).
\newblock \doi{10.1038/s41592-019-0686-2}

\bibitem{vostinar2024empirical}
Vostinar, A., Lalejini, A., Ofria, C., Dolson, E., Moreno, M.A.: Empirical: A
  scientific software library for research, education, and public engagement.
\newblock Journal of Open Source Software \textbf{9}(98), 6617 (2024).
\newblock \doi{10.21105/joss.06617}

\bibitem{waskom2021seaborn}
Waskom, M.L.: seaborn: statistical data visualization.
\newblock Journal of Open Source Software \textbf{6}(60), 3021 (2021).
\newblock \doi{10.21105/joss.03021}

\bibitem{yang2025downstream}
Yang, C., Wagner, J., Dolson, E., Zaman, L., Moreno, M.A.: Downstream:
  efficient cross-platform algorithms for fixed-capacity stream downsampling
  (2025).
\newblock \doi{10.48550/arXiv.2506.12975}

\bibitem{zaman2011rapid}
Zaman, L., Devangam, S., Ofria, C.: Rapid host-parasite coevolution drives the
  production and maintenance of diversity in digital organisms.
\newblock In: Proceedings of the 13th Annual Conference on Genetic and
  Evolutionary Computation (2011).
\newblock \doi{10.1145/2001576.2001607}

\bibitem{zhang2016cambricon}
Zhang, S., Du, Z., Zhang, L., Lan, H., Liu, S., Li, L., Guo, Q., Chen, T.,
  Chen, Y.: Cambricon-x: An accelerator for sparse neural networks.
\newblock In: 2016 49th Annual IEEE/ACM International Symposium on
  Microarchitecture (MICRO), pp. 1--12 (2016).
\newblock \doi{10.1109/micro.2016.7783723}

\bibitem{zhao2019elastic}
Zhao, X., Papagelis, M., An, A., Chen, B.X., Liu, J., Hu, Y.: Elastic bulk
  synchronous parallel model for distributed deep learning.
\newblock In: 2019 IEEE International Conference on Data Mining (ICDM) (2019).
\newblock \doi{10.1109/icdm.2019.00198}

\bibitem{zhu2025understanding}
Zhu, Z., Sun, Y., Parakal, D., Fang, B., Farrell, S., Bauer, G.H., Bode, B.,
  Foster, I.T., Papka, M.E., Gropp, W., Zhang, Z., Yang, L.: Understanding the
  landscape of ampere {GPU} memory errors (2025).
\newblock \doi{10.48550/arxiv.2508.03513}

\end{thebibliography}
 \newcommand{\noop}[1]{}

\endgroup
 
\printindex

\end{document}